\documentclass[10pt, a4paper, twocolumn, teaser, showabstract]{naverlabseurope}

\usepackage{lipsum}
\usepackage{multicol}
\usepackage{tikz,tkz-kiviat,pgfplots}

\usepackage{scalerel,xparse}
\NewDocumentCommand\emojifirst{}{
    \scalerel*{
        \includegraphics{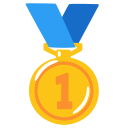}
    }{X}
}
\NewDocumentCommand\emojisecond{}{
    \scalerel*{
        \includegraphics{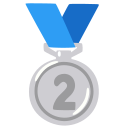}
    }{X}
}
\NewDocumentCommand\emojithird{}{
    \scalerel*{
        \includegraphics{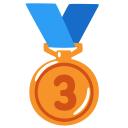}
    }{X}
}

\usepackage{color}
\usepackage{pifont}
\usepackage{multirow}

\newcommand{\mmhubert}{\texttt{mHuBERT-147}}
\newcommand{\blue}[1]{\textcolor{blue}{#1}}
\definecolor{green}{rgb}{0.0, 0.5, 0.0}
\newcommand{\green}[1]{\textcolor{green}{#1}}
\definecolor{orange}{rgb}{0.93, 0.53, 0.18}
\newcommand{\orange}[1]{\textcolor{orange}{#1}}
\newcommand{\red}[1]{\textcolor{red}{#1}}
\definecolor{purple}{rgb}{0.44, 0.16, 0.39}
\newcommand{\purple}[1]{\textcolor{purple}{#1}}

\usepackage{changes}
\definechangesauthor[name={Nikos}, color=orange]{nik}

\graphicspath{{figures/}}

\title{mHuBERT-147: A Compact Multilingual HuBERT Model}

\correspondingauthor{marcely.zanon-boito@naverlabs.com}

\authors{Marcely Zanon Boito¹, Vivek Iyer², Nikolaos Lagos¹, Laurent Besacier¹, Ioan Calapodescu¹}
\affiliations{¹NAVER LABS Europe - France\\²University of Edinburgh - UK}
\contributions{\textbf{Training code}: \url{https://github.com/utter-project/fairseq}\\\textbf{Pre-processing code}: \url{https://github.com/utter-project/mHuBERT-147-scripts}\\\textbf{Model Weights}: \url{https://huggingface.co/utter-project/mHuBERT-147}}

\teaserfig{\resizebox{\textwidth}{!}{
\begin{tabular}{lc|c|cc|c|ccc|c}
\toprule
\multicolumn{1}{c}{} & \textbf{}         & \textbf{Monolingual ASR} & \multicolumn{2}{c|}{\textbf{Multilingual ASR}} & \textbf{LID}  & \multicolumn{3}{c|}{\textbf{Multilingual ASR + LID}} & \textbf{}         \\
& & \textbf{normal} & \textbf{normal} & \textbf{few-shot} & \textbf{normal} & \textbf{normal} & \textbf{normal} & \textbf{few-shot} &\\
\multicolumn{1}{l}{\textbf{SSL}} & \textbf{\# Params} & \textbf{CER} ($\downarrow$)                                                    & \textbf{CER} ($\downarrow$)          & \textbf{CER} ($\downarrow$)          & \textbf{ACC} ($\uparrow$)  & \textbf{ACC} ($\uparrow$)   & \textbf{CER} ($\downarrow$)  & \textbf{CER} ($\downarrow$)  & \textbf{\textbf{SUPERB$_s$} ($\uparrow$)} \\\midrule
MMS-1B                  & 965M                & \textbf{33.3 / 25.7}                                                         & \textbf{21.3 / 18.1}           & \textbf{30.2 / 30.8}           & 84.8 / 86.1   & 73.3 / 74.8     & \textbf{26.0} / 25.5    & \textbf{25.4 / 24.8}    & \textbf{983.5}\emojifirst / 948.1\emojisecond \\\midrule
\textbf{mHuBERT-147}          & \textbf{95M}               & 34.2 / 26.3 & 23.6 / 22.0           & 33.2 / 32.9           & \textbf{85.3} / \textbf{91.0} &                \textbf{81.4} / 90.0	& 26.2 / 22.1 & 34.9 / 33.5  & 949.8\emojisecond / \textbf{950.2}\emojifirst \\
\midrule
MMS-300M                  & 317M              & 33.8 / 30.5                                                         & 28.7 / 24.0           & 36.5 / 36.5           & 62.3 / 84.3   & 71.9 / 74.3     & 31.5 / 30.0    & 30.9 / 29.2    &  824.9\emojithird / 844.3 \\
NWHC1                & 317M              & 39.5 / 30.5                                                         & 28.9 / 21.5           & 41.4 / 38.6           & 67.1 / 87.4 & 77.1 / \textbf{90.6}     & 28.8 / \textbf{21.5}    & 40.3 / 38.2    &   774.4 / 876.9\emojithird   \\
NWHC2                & 317M              & 39.5 / 30.5                                                         & 29.3 / 21.6           & 42.0 / 39.3           & 64.4 / 88.1   & 77.4 / \textbf{90.6}     & 28.4 / 21.8    & 41.5 / 38.8    &   759.9 / 873.3   \\
XLS-R-300M            & 317M              & 39.7 / 30.6                                                         & 29.2 / 22.0           & 40.9 / 39.3           & 66.9 / 87.9   & 55.6 / 85.6     & 28.4 / 22.9    & 42.1 / 42.4    &   730.8 / 850.5   \\
WavLabLM-large-MS    & 317M              & 40.5 / 32.8                                                         & 37.8 / 31.9           & 43.8 / 42.8           & 71.7 / 81.1   & 70.8 / 80.0     & 37.0 / 32.2    & 43.4 / 41.2    & 707.5 / 740.9 \\
\bottomrule
\end{tabular}}}
\teasercaption{ML-SUPERB 10min/1h leaderboard. Updated SOTA scores for each metric are presented in \textbf{bold}. The first~(\emojifirst), second~(\emojisecond) and third~(\emojithird) best SUPERB scores are highlighted.}

\begin{abstract}

We present \mmhubert{}, the first general-purpose massively multilingual HuBERT speech representation model trained on 90K hours of clean, open-license data. To scale up the multi-iteration HuBERT approach, we use faiss-based clustering, achieving 5.2x faster label assignment than the original method. We also apply a new multilingual batching up-sampling strategy, leveraging both language and dataset diversity. After 3 training iterations, our compact 95M parameter \mmhubert{} outperforms larger models trained on substantially more data. We rank second and first on the ML-SUPERB 10min and 1h leaderboards, with SOTA scores for 3 tasks. Across ASR/LID tasks, our model consistently surpasses XLS-R (300M params; 436K hours) and demonstrates strong competitiveness against the much larger MMS (1B params; 491K hours). Our findings indicate that \mmhubert{} is a promising model for multilingual speech tasks, offering an unprecedented balance between high performance and parameter efficiency.

\end{abstract}

\begin{document}

\maketitle

\section{Introduction}

Self-supervised Learning~(SSL) approaches for speech representation learning are the core foundation blocks of speech processing systems today. These models leverage large amounts of unlabeled speech data during training, and learn from different pretext tasks in order to build contextualized speech representations that can be leveraged for various downstream tasks. For English, several models have been proposed~\cite{pascual2019learning,paseplus,baevski2020wav2vec,hsu2021hubert,chen2022wavlm, liu2020mockingjay}, but most multilingual models available to the community are based on wav2vec~2.0~\cite{conneau21_interspeech,babu2021xls,pratap2023scaling}, with the only current exception being WavLabLM~\cite{chen2023joint}.
Meanwhile, the Hidden units BERT model~(HuBERT, \cite{hsu2021hubert}) -- where pre-training is performed in 2 or 3 iterations, and target training labels are externally obtained via k-means clustering -- presents superior performance on the English SSL benchmark SUPERB~\cite{yang2021superb}, outperforming wav2vec~2.0. The English version of this model also shows decent cross-lingual adaptation capabilities on the multilingual benchmark ML-SUPERB~\cite{mlsuperb_shi2023}. Recently, HuBERT has also emerged as a popular choice for producing discrete speech units for multimodal LLMs~\cite{wang2023viola,chang2023exploring,chou2023toward}.

However, despite recent efforts to reduce hardware costs~\cite{li2022improved,lin2023melhubert,chen2023reducing}, HuBERT's superior performance comes with higher training costs -- a characteristic that is further accentuated in multilingual settings requiring larger amounts of speech data. 
These cost demands arise from HuBERT's multi-iteration training process, which contrasts with wav2vec~2.0's single iteration training on unlabeled speech data alone. HuBERT requires high-dimensional feature extraction across the entire training dataset to generate discrete labels, along with a minimum of two model training and clustering steps, resulting in increased disk and CPU/GPU resource demands. 
We are aware of only small-scale multilingual HuBERT models that train using a single dataset: there is an mHuBERT trained on 3 languages~\cite{lee-etal-2022-textless}; and a HuBERT collection covering between 5 and 12 languages~\cite{duquenne-etal-2023-speechmatrix}.

In contrast, in this work, we tackle the challenge of training the first general-purpose massively multilingual HuBERT speech representation model. This model is trained using 90,430 hours of clean open-license data across 147 languages. For reducing pre-processing costs, we downsample large popularly used speech datasets, hypothesizing that source diversity is more important than quantity. We also propose to replace the original HuBERT clustering implementation with \textit{faiss}-based clustering~\cite{douze2024faiss}, increasing label assignment speed by 5.2 times. Finally, for training, we employ a two-level multilingual up-sampling approach factoring in both linguistic and dataset diversity for increased overall multilingual performance.

After three training iterations, and with only 95M parameters, our compact \mmhubert{} model outperforms multilingual wav2vec~2.0 models that are not only three times larger, but also trained with almost five times more data. It reaches respectively second and first place on the 10\,min and 1\,h leaderboard of the popular multilingual benchmark ML-SUPERB, with state-of-the-art (SOTA) scores for three out of four language identification~(LID) tasks. Across all ML-SUPERB tasks and settings, \mmhubert{} consistently surpasses XLS-R (300\,M parameters; 436\,K hours) -- its most comparable multilingual wav2vec~2.0 model. It also demonstrates strong competitiveness against the much larger MMS~(1\,B parameters; 491\,K hours). 
Complementary few-shot ASR evaluation on FLEURS-102 shows that our model is competitive to larger models, presenting impressive robustness at a compact size~(70\% less parameters).
Our findings suggest that \mmhubert{} is a promising compact model for speech processing pipelines, offering an unprecedented balance between high performance and parameter efficiency. We open-source all our scripts, manifest files and model weights.\footnote{See Appendix Section~\ref{appendix:section:opensource}.}

\section{\mmhubert{} data collection}
\label{sec:data}

\begin{figure*}
\centering
\resizebox{\textwidth}{!}{
\includegraphics{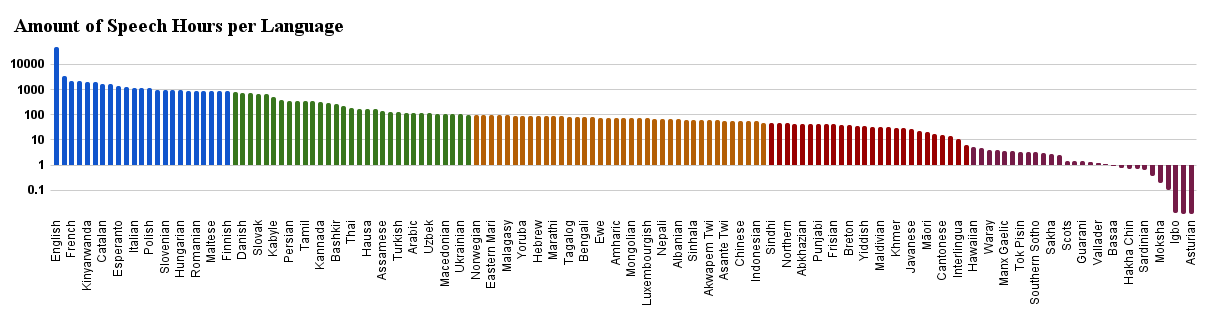}}
    \caption{Speech amount per language in a logarithmic scale, with different levels of speech resourcefulness: $\geq$~800\,h~(\blue{blue}), $\geq$~100\,h~(\green{green}), $\geq$~50\,h~(\orange{orange}), $\geq$~10\,h~(\red{red}), $\le$~10\,h~(\purple{purple}). Some languages are excluded from the plot. Best seen in color.}
\label{fig:data:langoverview}
\end{figure*}

We gathered 90,430 hours of speech from datasets with permissive licences in 147 languages. For this multilingual collection, our goal was to prioritize linguistic diversity over data quantity alone. 
Table~\ref{tab:licenses} lists the datasets we use along with their licences. Figure~\ref{fig:data:langoverview} illustrates data distribution across languages. In total, our training set spans 19 language families (sorted in decreasing order of data quantity): Indo-European, Niger-Congo, Uralic, Afro-Asiatic, Constructed~(Esperanto), Turkic, Dravidian, Sino-Tibetan, Austronesian, Koreanic, Kra-Dai, Japonic, Language isolate (Basque), Kartvelian, Austroasiatic, Mongolic, Northwest Caucasian, Creole and Tupian.
Appendix Tables~\ref{tab:languagefull:part1}-\ref{tab:languagefull:part5} list all included languages with amount of data per dataset.

Section~\ref{sec:data:datapreparation} details data pre-processing and filtering process.
Section~\ref{sec:data:overlap} discuss the overlap between the data we use and popular multilingual SSL models for speech representation learning.

%
\begin{table*}
\resizebox{\textwidth}{!}{
\begin{tabular}{llccc}
\toprule
\textbf{Dataset}                & \textbf{Full Names and References}      & \textbf{\# Languages}                                                                                                         & \textbf{\# Hours (filtered)} & \textbf{License}   \\\hline
\textbf{Aishells}                & Aishell~\cite{aishell_2017} and AISHELL-3~\cite{AISHELL-3_2020}   &   1                                                                                                               & 212                       & Apache License 2.0 \\\hline
\textbf{B-TTS}                  & BibleTTS~\cite{meyer2022bibletts}  &   6                                                                                                                   & 358                       & CC BY-SA 4.0       \\\hline
\textbf{Clovacall}              & ClovaCall~\cite{ha20_interspeech}   &   1                                                                                                                 & 38                        & MIT                \\\hline
\textbf{CV}                     & Common Voice version 11.0~\cite{commonvoice:2020}   & 98                                                                                                    & 14,943                    & CC BY-SA 3.0       \\\hline
\multirow{2}{*}{\textbf{G-TTS}} & \begin{tabular}[c]{@{}l@{}}High quality TTS data for Javanese, Khmer, \\ Nepali, Sundanese, and Bengali Languages~\cite{kjartansson-etal-tts-sltu2018}\end{tabular} & \multirow{2}{*}{9} & \multirow{2}{*}{27}       & \multirow{2}{*}{CC BY-SA 4.0}       \\
& \begin{tabular}[c]{@{}l@{}}High quality TTS data for four South\\ African languages~\cite{van-niekerk-etal-2017}\end{tabular}                             &  &                             &       \\\hline
\textbf{IISc-MILE}              & IISc-MILE Tamil and Kannada ASR Corpus~\cite{mile_1,mile_2}   &       2                                                                                & 406                       & CC BY 2.0          \\\hline
\textbf{JVS}                    & Japanese versatile speech~\cite{takamichi2019jvs}     &      1                                                                                            & 26                        & CC BY-SA 4.0       \\\hline
\textbf{Kokoro}                 & Kokoro Speech Dataset~\cite{kokorogithub}     & 1                                                                                                                   & 60                        & CC0                \\\hline
\textbf{kosp2e}                 & Korean Speech to English Translation Corpus~\cite{cho21b_interspeech}           &   1                                                                       & 191                       & CC0                \\\hline
\textbf{MLS}                    & Multilingual LibriSpeech~\cite{Pratap2020MLSAL}   & 8                                                                                                     & 50,687                    & CC BY 4.0          \\\hline
\textbf{MS}                     & MediaSpeech~\cite{mediaspeech2021}      &     1                                                                                                          & 10                        & CC BY 4.0          \\\hline
\textbf{Samrómur}               & Samrómur Unverified 22.07~\cite{hedstrom2022}   & 1                                                                                                   & 2,088                     & CC BY 4.0          \\\hline
\textbf{TH-data}               & THCHS-30~\cite{THCHS30_2015} and THUYG-20~\cite{THUGY20_2015,THUGY20_sre_2015}    &   2                                                                                                               & 46                        & Apache License 2.0 \\\hline
\textbf{VL}                     & VoxLingua107~\cite{valk2021slt}      &     107                                                                                                         & 5,844                     & CC BY 4.0          \\\hline
\textbf{VP}                     & VoxPopuli~\cite{wang-etal-2021-voxpopuli}      &  23                                                                                                               & 15,494                    & CC0               \\\bottomrule
\end{tabular}}
\caption{Datasets used for training \mmhubert{} with their corresponding abbreviation used throughout the paper~(left), amount of languages and hours (after filtering) and licenses. Downloading URLs are listed at Appendix Table~\ref{tab:url}.}
\label{tab:licenses}
\end{table*}

\subsection{Dataset pre-processing and filtering}\label{sec:data:datapreparation}

We follow the default HuBERT pre-processing guidelines~\cite{hsu2021hubert}.
For all datasets, the speech data is converted to 16-bit 16kHz WAV files, with volume reduction of $5\%$ applied for datasets in which excessive clipping is observed during conversion. Data is filtered to the interval of $[2,30]$s  and any utterances outside this window are discarded. This is with the exception of the following, where file concatenation is performed: B-TTS (Akwapen Twi, Asante Twi, Hausa, Yoruba), IISc-MILE (Kannada), and JVS (Japanese). In these cases, we performed concatenation due to the already limited amount of speech samples we had available for training in those languages.

For TTS/ASR/ST datasets, only the training set is used, validation and test sets are always excluded to prevent data contamination. The complete manifest list can be found in our HuggingFace repository.\footnote{Manifest files available at: \url{https://huggingface.co/utter-project/mHuBERT-147-base-3rd-iter/tree/main/manifest}}
We now detail dataset-specific pre-processing.\footnote{Pre-processing scripts available at: \url{https://github.com/utter-project/mHuBERT-147-scripts/tree/main/01_dataset_preproc}}

\begin{itemize}
    \item \textbf{VL:} 
    Manual inspection revealed that some language splits contained a significant amount of music, noise, and silence-only files. To improve the quality of the data used by our models, and to accurately estimate the amount of speech data per language, we filtered the dataset using the \texttt{inaSpeechSegmenter} tool~\cite{ddoukhanicassp2018}. 
    In total, we removed over 332K potentially noisy utterances~(249K music, 83K noise/silence). More details are given in Appendix Section~\ref{app.data.voxlingua}. 
    \item \textbf{VP:} For languages with ample representation in the multilingual collection~(i.e., over 1,000h of speech - namely German, English, Spanish, French, and Dutch), we exclusively use the 10K subset of VP~(2019 and 2020). For the other 18 languages, we utilize talks from 2017 to 2020 (100K VP split).
\end{itemize}

\subsection{Data overlap with existing multilingual models}\label{sec:data:overlap}
Table~\ref{tab:data:comparison1} shows the overlap in the training data of \mmhubert{} and the multilingual wav2vec~2.0~\cite{baevski2020wav2vec} models: XLSR-53~\cite{conneau21_interspeech}, XLS-R~\cite{babu2021xls} and MMS~\cite{pratap2023scaling}. Looking at the table, we see that the model most similar to ours in terms of dataset overlap is XLS-R, with only BABEL not included in our training. This dataset is not freely available, and it comprises speech data in 17 low-resource African and Asian languages: Assamese, Bengali, Cantonese, Cebuano, Georgian, Haitian, Kazakh, Lao, Northern Kurdish, Pashto, Swahili, Tagalog, Tamil, Tok Pisin, Turkish, Vietnamese and Zulu. Of these, the only language we did not find an openly available dataset for was Zulu. Thus, \mmhubert{} covers 127 of the 128 languages of XLS-R, while supporting 20 additional languages.

\begin{table}
\centering
\scriptsize
\resizebox{\columnwidth}{!}{
\begin{tabular}{cccccc}
\toprule
\multicolumn{1}{c}{\textbf{Dataset}} & \textbf{\# Langs} & \textbf{XLSR-53} & \textbf{XLS-R} & \textbf{MMS} & \textbf{\mmhubert{}}       \\\midrule
\textbf{BABEL}~\cite{gales2014speech}      & 17 & $$ \ding{51} $$    & $$ \ding{51} $$ & $$ \ding{51} $$& \\\hline
\textbf{CV  v2}    & 38 & $$ \ding{51} $$    & $$ \ding{51} $$ & $$ \ding{51} $$ & $$ \ding{51} $$ \\\hline
\textbf{CV v6}     & 60  &    & $$ \ding{51} $$  & $$ \ding{51} $$ & $$ \ding{51} $$ \\\hline
\textbf{CV v9}    &  89 &    &  & $$ \ding{51} $$& $$ \ding{51} $$ \\\hline
\textbf{CV v11}   & 98  &    &  && $$ \ding{51} $$\\\hline
\textbf{MLS}      & 8  & $$ \ding{51} $$    & $$ \ding{51} $$ & $$ \ding{51} $$& $$ \ding{51} $$\\\hline
\textbf{VP}      & 23   &  & $$ \ding{51} $$  & $$ \ding{51} $$& \textbf{down-sampled} \\\hline
\textbf{VL}       & 107  &    & $$ \ding{51} $$  & $$ \ding{51} $$& \textbf{down-sampled} \\\hline
\textbf{MMS-lab-U}~\cite{pratap2023scaling}  & 1,362 &    &  & $$ \ding{51} $$& \multicolumn{1}{l}{} \\\hline
\textbf{Aishells, B-TTS, Clovacall}     &  \multirow{4}{*}{23}   &    &   & & \multirow{3}{*}{$$ \ding{51} $$} \\
\textbf{G-TTS, IIScMILE, JVS}      &    &    &   & &  \\
\textbf{Kokoro, kosp2e, MS}       &   &    &   & &   \\
\textbf{Samrómur, TH-data}       &   &    &   & &   \\
\bottomrule    
\end{tabular}}
\caption{Datasets included in the training of the multilingual models most comparable to the \mmhubert{} model.
We highlight that although here we consider previous versions of CV as being entirely comprised on later versions, this is a simplification, and some files might have been removed due to speakers requests or down-votes.
}
\label{tab:data:comparison1}
\end{table}

\section{\mmhubert{} training}

Our training follows the multi-iteration pre-training objective of HuBERT~\cite{hsu2021hubert}. This approach leverages an external acoustic unit discovery model~(k-means clustering) to provide frame-level targets for masked span prediction pretraining. For optimization, two weighted cross-entropy losses are computed over masked~($L_m$) and unmasked~($L_u$) spans, defined as $L= \psi L_m + (1-\psi)L_u$. $L_m$ and $L_u$ can both be defined using the generic loss function in Equation~\ref{eq:hubertloss}.
Here, $f$ refers to the masked prediction model. $X'=[x_1,..,x_T]$ is the masked speech utterance of frame length $T$. $M \subset {\{1,..,T\}}$ are the masked indices in X' for masked loss~($L_m$), and the unmasked indices otherwise~($L_u$).
$Z = [z_i,..,z_T]$ are the discrete labels from clustering. 
Finally, $P_f$ produces the distribution over the target indices at each time-step $t$. The authors of HuBERT highlight that higher values of $\psi$ result in a model analogous to language modeling, where the prediction model $f$ is forced to learn both the unmasked segments and the long-range temporal structure of the speech data. 

\begin{equation}\label{eq:hubertloss}
    L(f;X',M,Z) = \sum_{t \in M} \log P_f(z_t | ~X', t)
\end{equation}

Next, we dive into how \mmhubert{} differs from HuBERT training, including a) a more complex data up-sampling strategy~(\ref{sec:sampling}), and b) a replacement of the original k-means implementation with the efficient \textit{faiss}~\cite{douze2024faiss} Inverted File Index~(IVF) -- drastically increasing labeling speed~(\ref{sec:faiss}). Finally, we discuss our experimental setup~(\ref{sec:experimentalsetting}).

\subsection{Two-level language-source up-sampling}\label{sec:sampling}

To optimize for exposure to different languages and data sources during training, we employ a two-level up-sampling strategy during multilingual batching.
Let $N$ be the total number of examples in the training set, with $n_{l}$ corresponding to the count for language~$l$.
The sampling probability for $l$ is $P_{l} \propto \left(\frac{n_l}{N}\right)^{\alpha}$, where $\alpha$ is a hyper-parameter in $[0,1]$; with $\alpha = 1$ resulting in no up-sampling, and lower values resulting in higher probabilities for under-represented languages.

For each epoch, we sample N times from the probability distributions $P_l$.
In this way we reach a quantity $B_l$ of examples selected per language $l$. 
We sample $B_l$ utterances by considering varied data sources~(datasets). The probability of sampling an utterance of $l$ from data source $x$ is given by $P_{x} \propto \left(\frac{n_l(x)}{n_l}\right)^{\beta}$, where $n_l(x)$ corresponds to the number of examples of language $l$ from data source $x$, and $\beta$ is a hyper-parameter in $[0,1]$. 
We sort the $N$ selected utterances by length before batching, to minimize random cropping.

The most similar up-sampling strategy to ours is MMS -- which considers (language, dataset) pairs as unique languages for up-sampling. However, their approach inflates the training time allocated to high-resource languages, as these tend to have more diverse data sources in the training set. Unlike them, we adopt a two-level technique to better suit our diverse dataset collection.

\subsection{Efficient clustering with faiss indices}\label{sec:faiss}
We use \textit{faiss}-based~\cite{douze2024faiss} k-means clustering for faster label assignment. 
This library facilitates efficient similarity search and clustering of dense vectors. We cluster using Inverted File Index~(IVF) with the following setting: \texttt{OPQM\_D,IVFK\_HNSW32,PQMx4fsr}, that we now detail.\footnote{More information can be found at: \url{https://github.com/facebookresearch/faiss/wiki/The-index-factory}}
\begin{itemize}
    \item 
``\texttt{OPQM\_D,..,PQMx4fsr}'' is used for indexing RAM usage. This option optimizes vector representation by rotation and then performs input vector projection into dimension D. Product quantization~(PQ) is then applied to hash the vectors into M 4-bit codes, resulting in the storage of M/2 bytes per vector. We use $M=16,D=64$.
    \item 
    ``\texttt{..,IVFK\_HNSW32,..}'' denotes the indexing itself. IVF performs coarse quantization via an efficient implementation of k-means. This option uses Hierarchical Navigable Small Worlds~(HNSW) graphs for cluster assignment, with 32 being the number of links per vertex. In practice, this greatly speeds up clustering (5.2x faster than \citet{hsu2021hubert}).
\end{itemize}

\subsubsection{Faiss vs sklearn k-means}\label{section:3:faissvssklearn}
Our motivation to replace the \texttt{sklearn} k-means by \texttt{faiss} comes from our struggle to scale this approach to a multilingual setting with more data and a higher K. For instance, working with a 850\,GB set of vectors of dimensionality 768, we were unable to finish the clustering step after four days, while \texttt{faiss} implementation was able to produce an index in 40\,h. Moreover, label application is 5.2 times faster by using \texttt{faiss}  indices compared to \texttt{sklearn}  clustering. This speed up in label application is particularly relevant for our work, as we have a massive training set.

We verified that this new clustering setting does not impact downstream model quality by pretraining two baselines: a two-iteration English HuBERT following the original codebase, and a \texttt{faiss}-based equivalent model with the same number of centroids. Both pretrained models were then fine-tuned on the Librispeech~\cite{7178964} 100\,h ASR, performing similarly on clean/other test sets. 
Appendix Section~\ref{appendx:sec:faissvssklearn} presents our detailed results and discussion.

\subsection{Experimental setting}\label{sec:experimentalsetting}

\subsubsection{Faiss training}
For clustering, we also sample data using the two-level up-sampling approach, to ensure alignment with the training data. We maximize the data quantity used for clustering by sampling with a RAM budget of 850\,GB, which corresponds to over 6\,M examples~(20.8\%) for iteration 1, and over 630\,K examples~(around 2\%) for iteration 2 and 3. The reduced example count for later iterations is due to the features being high-dimensional~($dim=768$) compared to iteration 1 MFCCs~($dim=39$). We use $K=1000$ across all iterations.\footnote{Code available at: \url{https://github.com/utter-project/mHuBERT-147-scripts/tree/main/03_faiss_indices/}}

\subsubsection{Pre-training setting}
We build our codebase on top of \textit{fairseq}~\cite{ott-etal-2019-fairseq}, introducing data loading optimizations and a new multilingual batching setting~(Section~\ref{sec:sampling}).\footnote{Fork available at: \url{https://github.com/utter-project/fairseq}} We train HuBERT base models (95\,M parameters) following the original recipe's hyperparameters but with a larger batch size (2.8\,M instead of 1.4\,M frames), input normalization, and scaling to more updates: 2\,M steps (approximately 31 epochs), instead of the original 400\,K. For our two-level up-sampling strategy, we select $\alpha=0.7$~(language up-sampling); and $\beta=0.9$.~(source up-sampling). 
Feature extraction for iteration 2 and 3 is performed using respectively the 6th and the 9th layer of the previous iteration last checkpoint.  
Training one iteration requires 32x~A100-80GB GPUs running in 4 parallel nodes for approximately 20 days. More details regarding training are presented in Appendix Section~\ref{appendix:pretraining:settings}.

\subsubsection{Hardware considerations}

\mmhubert{} is trained on over 90\,K hours of speech in 147 languages -- considerably lesser than its most comparable multilingual model XLS-R~(128 languages, 436\,K\,hours), and the larger scale MMS~(1,406 languages, 491\,K\,hours).
Indeed, data quantity is a major bottleneck for HuBERT, since the training process requires discrete labels for the entire training dataset which, in turn, are produced from high-dimensional features. Extracting these features in a typical multi-iteration process can incur prohibitive expenses for massive datasets, in terms of both storage requirements and pre-processing times. For instance, the features we extracted for \mmhubert{} occupied 48\,TB of storage per iteration. We estimate that for the training set of XLS-R, this would take about 270\,TB. While storage requirements can be reduced by looping the feature extraction and labeling processes, this would increase pre-processing times. Appendix Section~\ref{appendix:sec:hardware} details hardware requirements for the different pre-processing and training steps.
\section{ML-SUPERB evaluation}\label{sec:mlsuperb}

\begin{table*}[h!]
\scriptsize
\resizebox{\textwidth}{!}{
\begin{tabular}{lc|c|cc|c|ccc|c}
\toprule
\multicolumn{1}{c}{} & \textbf{}         & \textbf{Monolingual ASR} & \multicolumn{2}{c|}{\textbf{Multilingual ASR}} & \textbf{LID}  & \multicolumn{3}{c|}{\textbf{Multilingual ASR + LID}} & \textbf{}         \\
& & \textbf{normal} & \textbf{normal} & \textbf{few-shot} & \textbf{normal} & \textbf{normal} & \textbf{normal} & \textbf{few-shot} &\\
\multicolumn{1}{l}{\textbf{SSL}} & \textbf{\# Params} & \textbf{CER} ($\downarrow$)                                                    & \textbf{CER} ($\downarrow$)          & \textbf{CER} ($\downarrow$)          & \textbf{ACC} ($\uparrow$)  & \textbf{ACC} ($\uparrow$)   & \textbf{CER} ($\downarrow$)  & \textbf{CER} ($\downarrow$)  & \textbf{\textbf{SUPERB$_s$} ($\uparrow$)} \\\midrule
MMS-1B                  & 965M                & \textbf{33.3 / 25.7}                                                         & \textbf{21.3 / 18.1}           & \textbf{30.2 / 30.8}           & 84.8 / 86.1   & 73.3 / 74.8     & \textbf{26.0} / 25.5    & \textbf{25.4 / 24.8}    & \textbf{983.5}\emojifirst / 948.1\emojisecond \\
NWHC1                & 317M              & 39.5 / 30.5                                                         & 28.9 / 21.5           & 41.4 / 38.6           & 67.1 / 87.4 & 77.1 / \textbf{90.6}     & 28.8 / \textbf{21.5}    & 40.3 / 38.2    &   774.4 / 876.9\emojithird   \\
NWHC2                & 317M              & 39.5 / 30.5                                                         & 29.3 / 21.6           & 42.0 / 39.3           & 64.4 / 88.1   & 77.4 / \textbf{90.6}     & 28.4 / 21.8    & 41.5 / 38.8    &   759.9 / 873.3   \\\midrule
\textbf{mHuBERT-147-3rd}          & \textbf{95M}               & 34.2 / 26.3 & 23.6 / 22.0           & 33.2 / 32.9           & \textbf{85.3} / \textbf{91.0} &                \textbf{81.4} / 90.0	& 26.2 / 22.1 & 34.9 / 33.5  & 949.8\emojisecond / \textbf{950.2}\emojifirst \\
\textbf{mHuBERT-147-2nd}          & \textbf{95M}               & 35.9 / 27.6 & 25.4 / 22.5           & 34.2 / 33.8           & 74.8 / \textbf{90.1} &                \textbf{81.0} / 89.0	& 26.3 / 23.6 & 33.9 / 34.4  & 895.0 / 925.7 \\
\midrule
MMS-300M                  & 317M              & 33.8 / 30.5                                                         & 28.7 / 24.0           & 36.5 / 36.5           & 62.3 / 84.3   & 71.9 / 74.3     & 31.5 / 30.0    & 30.9 / 29.2    &  824.9\emojithird / 844.3 \\
XLS-R-300M            & 317M              & 39.7 / 30.6                                                         & 29.2 / 22.0           & 40.9 / 39.3           & 66.9 / 87.9   & 55.6 / 85.6     & 28.4 / 22.9    & 42.1 / 42.4    &   730.8 / 850.5   \\
WavLabLM-large-MS    & 317M              & 40.5 / 32.8                                                         & 37.8 / 31.9           & 43.8 / 42.8           & 71.7 / 81.1   & 70.8 / 80.0     & 37.0 / 32.2    & 43.4 / 41.2    & 707.5 / 740.9 \\
\bottomrule
\end{tabular}}
\caption{ML-SUPERB 10min/1h results. Current SOTA~\cite{shi2023findings} is shown on the top portion of the table, our submission is shown in the middle, other relevant multilingual models are presented below it. Updated SOTA scores for each metric are presented in \textbf{bold}. The first~(\emojifirst), second~(\emojisecond) and third~(\emojithird) best SUPERB scores are highlighted.}
\label{tab:mlsuperb}
\end{table*}

For evaluating the quality of the multilingual representations learned by \mmhubert{}, we use the ML-SUPERB benchmark~\cite{mlsuperb_shi2023}. This benchmark comprises two settings: 10min and 1h; and four tasks: monolingual ASR; multilingual ASR, LID, and joint multilingual ASR and LID. 
The monolingual setting constitutes 13 (language, domain) pairs,
while the multilingual one has 240 pairs across 143 languages in total -- of which 123 and 20 constitute the \textit{normal}~(10min/1h per language) and \textit{few-shot}~(5 utterances per language) evaluation settings respectively.

\paragraph{Setup.} The downstream architecture consists of learnable weights over the frozen SSL features, a CNN for reducing feature dimensionality by a half, and two Transformer layers~($dim=256$; 8 attention heads). In line with the official guidelines,\footnote{Recipe at: \url{https://github.com/espnet/espnet/tree/master/egs2/ml_superb/asr1}}  we only tune the learning rates~(best of $1e-3;1e-4;1e-5$) on the validation set.\footnote{Appendix~\ref{appendix:experiments:mlsuperb} presents more information regarding optimization.} Due to the high compute costs of evaluating on this benchmark, we do not retrain the other submissions we compare against and instead reuse the leaderboard scores from~\citet{shi2023findings}. Therefore, we are unfortunately unable to provide confidence intervals, as these were not originally requested by the benchmark organizers. 
We also recompute the global SUPERB scores for all models following~\citet{mlsuperb_shi2023}, since this metric leverages SOTA scores for normalization, and our model achieves new SOTA on three tasks~(bold values in Table~\ref{tab:mlsuperb}).\footnote{Our SUPERB score calculator is available at: \url{https://github.com/utter-project/mHuBERT-147-scripts/blob/main/06_evaluation/mlsuperb/compute_superb_score.py}}

\paragraph{Results.} Table~\ref{tab:mlsuperb} presents results for the 10min/1h settings. The current SOTA models from~\citet{shi2023findings} are shown at the top~(NWHC models are variants of MMS-300M). Other relevant multilingual SSL models are displayed in the bottom portion: MMS-300M; XLS-R-300M, and the best WavLabLM model from~\citet{chen2023joint}~(136 languages; 40\,K hours). We highlight that although XLS-R and MMS are referred in the literature as ``300\,M'', the correct parameter count is 317\,M.
We omit XLSR-53 results as these are consistently worse than XLS-R~\cite{mlsuperb_shi2023}. We present results for both our second~(2nd) and third~(3rd) \mmhubert{} iterations. 
\\Looking at the bottom rows, we notice that \mmhubert{} outperforms the XLS-R and WavLabLM models across all tasks starting from its 2nd iteration. Compared to MMS-300M, both 2nd and 3rd iterations of our model are only surpassed in three tasks -- few-shot ASR+LID 10min/1h, and monolingual ASR 10min -- and by a small margin. These results highlight that the \mmhubert{} training approach produces effective output starting from the 2nd iteration. They also motivate the interest in conducting a 3rd iteration, and we observe further improvement over all tasks between \mmhubert{}\texttt{-2nd} and \mmhubert{}\texttt{-3rd} scores.
\\Focusing on our best \mmhubert{} model, and comparing it against the current SOTA~(top rows), we see that this model is again very competitive, despite being trained on much lesser data. 
Our compact \mmhubert{}\texttt{-3rd} reaches the first position of the leaderboard for the 10min setting, and the second position for the 1h setting, being outperformed only by a model ten times larger~(MMS-1B).
We also set new SOTA ACC scores for LID~10min, LID~1h and Multilingual ASR+LID 10min, while being the only 95\,M parameters model at the top of the leaderboard.

\section{FLEURS-102 evaluation}\label{sec:fleurs102}

We complement the ML-SUPERB evaluation by training monolingual ASR models on the FLEURS-102 dataset~\cite{conneau2023fleurs}, competing with XLS-R and MMS~(300M/1B). In this full fine-tuning few-shot setting, we simply add a linear projection to the target vocabulary on top of the pre-trained stack, optimizing the resulting model for ASR using approximately 10 hours of speech. Thus, unlike the experiments in the previous section, larger models will now have the benefit of having more parameters for adaptation. With these experiments, we want to illustrate that even in this unfavorable setting, \mmhubert{} can still be competitive, while being faster at fine-tuning and inference time.

\paragraph{Setup.} We implement monolingual ASR using the \textit{transformers} library~\cite{wolf-etal-2020-transformers}. All models were trained for 100 epochs on \textit{fp32} 
using V100-32GB/A100-80GB GPUs. Since individual language optimization would be prohibitively costly for this dataset, we select a subset of 29 languages, covering the different geographic groups, and optimize parameters using XLS-R. We use $1e-5$ as learning rate, warm-up ratio of $0.1$ and dropout of $0.1$~(300\,M and 1\,B) or $0.3$~(95\,M). The increased dropout for the latter is due to the ASR models being considerably smaller~(70\% less parameters). We train three models per language with different seeds, adding up to four runs in cases of high variability in scores~($\geq20$~CER). We apply MMS transcript normalization~\cite{pratap2023scaling}, reporting CER averages over the two best runs.

\paragraph{Results.} Table~\ref{tab:fleurs} presents results grouped by FLEURS-102 geographic groups: Western Europe~(WE); Eastern Europe~(EE); Central-Asia/Middle-East/North-Africa~(CMN); Sub-Saharan Africa~(SSA); South-Asia~(SA); South-East Asia~(SEA); Isolates~(CJK). 
Despite its small size, \mmhubert{} achieves the best average over 102 languages in this setting, mainly due to its superior performance in the CMN, SA, and SEA groups. Further investigation reveals this superior performance is due to its greater robustness to few-shot adaptation compared to other models. For instance, compared to MMS-1B~(See Appendix Table~\ref{appendix:tab:fleurs102}), \mmhubert{} consistently shows higher CER averages, likely due to its smaller capacity. However, MMS-1B fails to achieve useful transcription (CER$>90$) for 16 languages, while \mmhubert{} only fails for half of these.\footnote{Removing the aforementioned 16 languages from the computed average, we find that MMS-1B indeed produces the best FLEURS-102 scores~(MMS-1B: 8.2, \mmhubert{}: 13.3).} In other words, its overall superior performance is due to its representations converging more consistently in few-shot settings compared to other models. Appendix Section~\ref{appendix:experiments:fleurs102} present more information regarding the FLEURS-102 results.

\begin{table}
\centering
\scriptsize
\resizebox{\columnwidth}{!}{
\begin{tabular}{lc|ccccccc}\toprule
\textbf{SSL}         & \textbf{\begin{tabular}[c]{@{}c@{}}General\\ Avg (102)\end{tabular}} & \textbf{\begin{tabular}[c]{@{}c@{}}WE\\ (25)\end{tabular}} & \textbf{\begin{tabular}[c]{@{}c@{}}EE\\ (16)\end{tabular}} & \textbf{\begin{tabular}[c]{@{}c@{}}CMN\\ (12)\end{tabular}} & \textbf{\begin{tabular}[c]{@{}c@{}}SSA\\ (20)\end{tabular}} & \textbf{\begin{tabular}[c]{@{}c@{}}SA\\ (14)\end{tabular}} & \textbf{\begin{tabular}[c]{@{}c@{}}SEA\\ (11)\end{tabular}} & \textbf{\begin{tabular}[c]{@{}c@{}}CJK\\ (4)\end{tabular}} \\\midrule
\textbf{MMS-1B}      & 22.3 & \textbf{17.4} & \textbf{11.0} & 37.8 & 23.3 & 27.7 & 25.9 & \textbf{17.8} \\
\textbf{MMS-300M}    & 24.9 & 19.5 & 12.5 & 39.8 & 24.8 & 29.1 & 29.5 & 35.8 \\
\textbf{XLS-R-300M}  & 24.5 & 18.7 & 11.8 & 39.2 & 24.8 & 29.7 & 29.5 & 33.4 \\
\textbf{\mmhubert{}} & \textbf{21.1} & 18.7 & 15.3 & \textbf{23.4} & \textbf{22.7} & \textbf{25.5} & \textbf{19.8} & 31.5  \\\bottomrule
\end{tabular}       
}
\caption{FLEURS-102 CER ($\downarrow$) geographic group averages, with number of languages between parentheses.}
\label{tab:fleurs}
\end{table}

\paragraph{Training and Inference efficiency.} We measured fine-tuning~(1 epoch) and inference efficiency~(test set) across three runs and two languages~(English and Kannada) using a RTX~3090-24GB in exclusive node execution mode. We find that \mmhubert{}, which has 70\% less parameters, is respectively 1.8 and 3 times faster to fine-tune on average than 300M and 1B models. For test inference, our model is respectively 1.4 and 2 times faster on average than 300M and 1B models.

Overall, our results highlight our model as a compact but powerful solution for multilingual speech processing applications. Appendix Section~\ref{appendix:experiments:monolingual} present complementary results for the ESB benchmark.

\section{Conclusion}

In this paper we presented the first general-purpose multilingual HuBERT model. For training, we prioritized data quality over quantity -- selecting diverse data sources and filtering noisy data, and defining a two-level multilingual batching up-sampling approach that considers both language and dataset diversity. For clustering, we leveraged \textit{faiss}-optimized clustering for decreasing the label assignment costs.
Despite being trained with considerably less data than popular multilingual models, and being a compact model, \mmhubert{} is competitive with larger multilingual SSL models, reaching respectively second and first positions at ML-SUPERB 10min/1h leaderboards, and setting new SOTA for three LID metrics. Complementary ASR results on FLEURS-102 suggest that \mmhubert{} is a promising model for multilingual speech downstream tasks, offering an unprecedented balance between high performance and parameter efficiency.

\clearpage

\section{Acknowledgments}
We thank the ML-SUPERB authors for providing us with assistance and clarification during our models evaluation. We would also like to thank our NAVER LABS Europe colleagues Caroline Brun and Salah Aït-Mokhtar for their assistance with the FLEURS-102 experiments.

This work was funded by the European Union’s Horizon Europe~(HE) Research and Innovation programme under Grant Agreement No 101070631 and from the UK Research and Innovation (UKRI) under the UK government's HE funding grant No 10039436. This document is licensed under CC-BY-4.0.
\vskip 0.2cm

\centerline{\includegraphics[height=1.3cm]{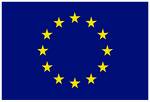}
\hfill
\includegraphics[height=1.3cm]{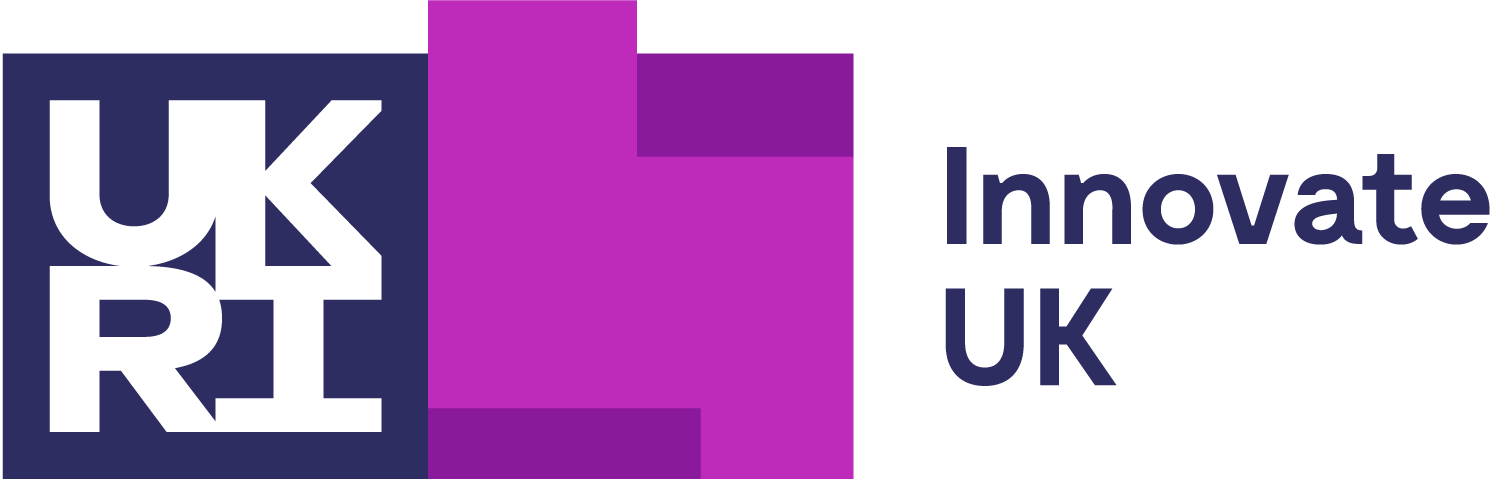}}

{
    \small
    \bibliographystyle{ieeenat_fullname}
    \bibliography{main}
}

\clearpage
\appendix
\clearpage
\section{Pre-training experiments}
\label{app.experiments}

\subsection{Faiss-based clustering}\label{appendx:sec:faissvssklearn}

For evaluating our hypothesis that the mini-batch k-means proposed in the original HuBERT recipe can be replaced with the more efficient \texttt{faiss} IVF clustering (refer Section~\ref{sec:faiss}), 
we trained two HuBERT base models~\cite{hsu2021hubert} from scratch - one with k-means clustering and the other with \texttt{faiss} indices. We follow the recipe available in the \texttt{fairseq} library~\cite{ott-etal-2019-fairseq}\footnote{Available at: \url{https://github.com/facebookresearch/fairseq/blob/main/examples/hubert/README.md}} for training the HuBERT models.  The models are trained for two iterations on the LibriSpeech dataset~\cite{panayotov2015librispeech}. We use the same number of centroids as the original recipe -- $K=100$ in the first iteration, and $K=500$ in the second.\footnote{Model hyper-parameters are available at: \url{https://github.com/facebookresearch/fairseq/blob/main/examples/hubert/config/pretrain/hubert_base_librispeech.yaml}}

\paragraph{Clustering performance.}
We find that, in this setting, both approaches take the same time to train their labels when using Librispeech train split with $K=500$ in the second iteration~(2.33 hours). However, as mentioned previously (Section~\ref{section:3:faissvssklearn}), after four days, we failed to finish training a \texttt{sklearn}-based k-means using more data~(850\,GB) and higher K~($K=1000$). For \texttt{faiss}, the same setting took approximately 40 hours. 

\paragraph{Labeling performance.} Clustering application using \texttt{sklearn} takes 1.8\,min per 10\,h of speech, whereas it takes \texttt{faiss} only 21\,s to label the same amount of speech data. This makes the former approach 5.2 times slower. The speed up in the latter is mainly due to the Hierarchical Navigable Small Worlds~(HNSW) optimized graph search algorithm.

\paragraph{LibriSpeech Results.} We train models on the 100\,h split from LibriSpeech, using the Hugging Face \texttt{transformers} library~\cite{wolf-etal-2020-transformers}. We train for 30 epochs on \texttt{fp16} precision, adding a linear projection to the target vocabulary on top of the pre-trained stacks. We explore two fine-tuning approaches: \textit{all}~(wherein all weights are updated) or \textit{weighted}~(wherein only downstream layers and a weighted sum over the pre-trained representations are updated). We tune the learning rate for each model~(best of $1e-2;1e-3;1e-4;1e-5$), 
selecting the best result using valid-clean. 
We train models twice, varying the random seed, and present average results.
Table~\ref{tab:appendix:faiss:lsvalid} presents results for the original HuBERT model\footnote{Available at: \url{https://huggingface.co/facebook/hubert-base-ls960}}~(original), our reproduction of the same model~(sklearn), and the faiss-based clustering version~(faiss). 
\\We observe that both models trained from scratch~(sklearn and faiss) present a slight decrease in performance compared to the original pre-trained model. We believe this could be due to differences in data selection for the clustering steps, which is performed at random using 20\% of the training data. Our models are also trained using the random crop batching approach, while the original implementation might have used padding.\footnote{The shared recipe code mentions that ``v1'' HuBERT used padding, but we could not find information in the paper regarding this.}
Finally, focusing on models trained under the same settings, we observe that performance for sklearn and faiss models is very similar. This suggests that faiss IVFs can be leveraged for HuBERT training without performance degradation. In Appendix Section~\ref{sec:app:faissvskmeansmultilingual} we illustrate that in multilingual settings, and by increasing K for \texttt{faiss}, we also produce equally performing speech representation models by using either \texttt{sklearn} or \texttt{faiss} clustering.

\begin{table}
\resizebox{\columnwidth}{!}{
\begin{tabular}{lccc}
\toprule
         & \textbf{FT}               & \textbf{test-clean ($\downarrow$)}  & \textbf{test-other ($\downarrow$)}  \\\midrule
\texttt{original} & \multirow{3}{*}{all}      & 7.1                  & 15.3                 \\
\texttt{sklearn}  &                           & 8.4 & 17.1 \\
\texttt{faiss}    &                           & 7.9                  & 17.1                 \\\midrule
\texttt{original} & \multirow{3}{*}{weighted} & 48.8                 & 57.1                 \\
\texttt{sklearn}  &                           & 50.5 & 59.0 \\
\texttt{faiss}    &                           & 50.2                 & 59.2            \\\bottomrule    
\end{tabular}}
\caption{Average WER test results for the Librispeech dataset using different English 2nd iteration HuBERT models.}
\label{tab:appendix:faiss:lsvalid}
\end{table}

\subsection{Hardware considerations}\label{appendix:sec:hardware}

In this section, we detail the different hardware requirements for training our \mmhubert{} model. Table~\ref{tab:hardwareinfo}  summarizes the key information. The next few paragraphs elaborate on these costs.

\begin{table*}
\centering
\resizebox{\textwidth}{!}{
\begin{tabular}{ll|c|c|c|c|c}\toprule
  & \textbf{Task}                    & \textbf{RAM/job} & \textbf{CPU/job}                                                                & \textbf{GPU/job} & \textbf{Total Disk} & \textbf{\begin{tabular}[c]{@{}c@{}}Total Processing \\ Time\end{tabular}} \\\midrule
0 & Speech                   & -                & -                                                                               & -                & 9.6 TB              & -                                                                         \\
1 & Speech Features iteration 1       & -                & -                                                                               & -                & 4.7 TB              & -                                                                         \\
1 & Speech Features iteration 2-3 & 50-500 GB        & -                                                                               & 1x V100-32 GB   & 48 TB               & 3-5 days                                                                  \\
2 & \texttt{faiss} Index Training             & 2 TB             & \begin{tabular}[c]{@{}c@{}}32x Intel(R) Xeon(R) \\ Platinum 8452Y\end{tabular}  & -                & 3.6 GB              & 2 days                                                                    \\
3 & \texttt{faiss} Index Application          & 50-500 GB        & 4-8 cores (any)                                                                 & -                & 37 GB               & 2-3 days                                                                  \\
4 & Manifest/Labels                  & -                & -                                                                               & -                & 65 GB               & -                                                                         \\
5 & Model training (iteration)       & 2 TB/node        & \begin{tabular}[c]{@{}c@{}}64 x AMD EPYC \\ 7313 16-Core Processor\end{tabular} & 32x A100-80GB     & 35 GB               & 20 days                         \\\bottomrule                                         
\end{tabular}}
\caption{Overview of the hardware requirements necessary per job~(RAM; CPU; GPU), total disk required to store the output of the tasks, and total estimated processing time. The numbers in the left illustrate the hierarchy of processes, with larger numbers depending on the completion of the previous processes.}
\label{tab:hardwareinfo}
\end{table*}

\paragraph{\textbf{Data storage requirements.}}
For storing \mmhubert{}'s 90\,K hours of speech, 9.6\,TB of disk storage are required.
For the first iteration, an additional 4.7\,TB is required for storing the extracted MFCC features for clustering and label application. For the second and third iterations, substantially more storage is needed -- as the 39-dimensional features are replaced by 768-dimensional vectors. As a result, our storage requirements increased roughly 10 times, reaching about 48\,TB. Fortunately, the storage of feature vectors is not a requirement for training, and so in our pre-processing protocol, we performed iterative extraction for labeling, avoiding having more than 10\,TB of features vectors at a given moment on our clusters. 
Thus, we can consider that the upper bound of storage required to train a \mmhubert{} model with 90\,K hours is around 58\,TB, and we can thus estimate that training a model with the same amount of speech data as XLS-R could necessitate up to 280\,TB of storage space. While one can cut storage costs during pre-processing, this comes with an increase in pre-processing time. The iteration between extraction and labeling, followed by feature deletion, might also extend the duration of GPU idle time between consecutive iterations.

\paragraph{\textbf{GPU requirements for pre-processing.}} GPUs are required to extract features for the entire corpus, as these are needed to produce the target labels. The cost of this extraction can be greatly reduced with data sharding. On average, 1\,h of speech vectors requires 9\,s in a V100-32GB. Considering 12 parallel jobs running in V100-32GB GPU, with optimum sharding 90\,K hours of speech data could be processed in approximately 19\,h. However, this is an over-optimistic lower-bound, as sharding will never be optimal unless it is performed cross-dataset~(mixing data from different datasets during extraction process), which makes data management a more complex task. Moreover, this also requires the upper-bound of storage requirements discussed above.
For \mmhubert{}, we shard feature extraction only for datasets containing over 20,000 utterances for a given language, so as to allow us to more effectively parallelize the labeling task. This results in an approximate extraction time of 3 to 5 days for the entire \mmhubert{} training corpus.

\paragraph{\textbf{Clustering requirements.}} 
In the original HuBERT paper, this step is performed by clustering with mini-batch k-means. In Section~\ref{sec:faiss} we discuss replacing this approach with \texttt{faiss} indices, that are much more scalable.  We sample training data using the same distribution used for up-sampling during multilingual training~(Section~\ref{sec:sampling}), and by having a fixed RAM target of around 850\,GB, which allows us to train indices in 2\,TB RAM machines.\footnote{The \texttt{faiss} index creation implementation requires almost twice the amount of RAM used for loading the data in order to train its indices.} Index training using the ``\texttt{OPQ16\_64,IVF1000\_HNSW32,PQ16x4fsr}'' option takes approximately 40\,h running on exclusivity mode on a node with 32 Intel(R) Xeon(R) Platinum 8452Y and 2\,TB RAM. Index application~(i.e. labeling of speech feature vectors) takes 21\,s per 10\,h of speech transformed into \texttt{numpy} feature vectors, costing about 53.65 processing hours for the totality of the training data. Index application is a RAM-bound process, where loading of the \texttt{numpy} features occupies the majority of the processing time, and thus, with increased sharding and parallel processing, index application takes negligible processing time.

\paragraph{\textbf{Training requirements.}} Training was performed using the NAVER Cloud platform.\footnote{\url{https://www.ncloud.com/}} We use 32x~A100-80\,GB GPUs running on 4 parallel nodes. Regarding RAM requirements, considering 8 parallel data loaders, manifest lists and label offsets stored in memory, 1.5\,TB of RAM is needed. While keeping labels directly in RAM would increase batching speed, this would require an additional 2\,TB of RAM. Storage needs are limited to the speech data (9.6\,TB), corresponding manifest and label files~(65\,GB) and checkpoints~(1.5\,GB per checkpoint).

\subsection{Loss curves and training instability}
\begin{figure*}
\centering
\resizebox{\textwidth}{!}{
\includegraphics{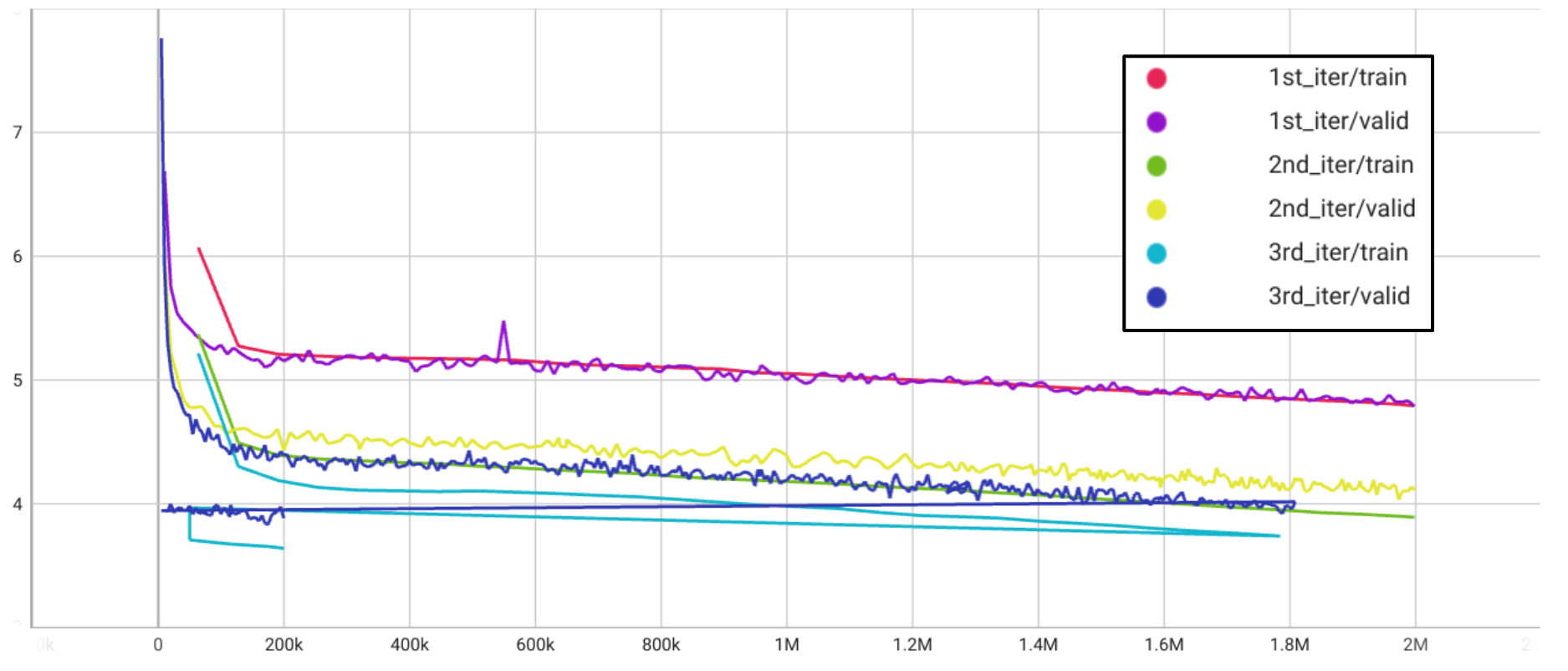}}
    \caption{Loss curves for the 3 iterations of \mmhubert{} training. For the 3rd iteration, the step jump from 1.8M to 0 is due to the optimizer re-initialization: this model crashed in \texttt{fp16} and had to be reinitialized using \texttt{fp32} for the last 200\,K updates. Best seen in color.}
\label{fig:training:losscurves}
\end{figure*}

The loss curves for the three \mmhubert{} iterations are presented in Figure~\ref{fig:training:losscurves}. We observe that while the loss decrease from the 1st to the 2nd iteration is very pronounced, it seems to be more limited between the 2nd and the 3rd iterations. We hypothesize that the model might be starting to saturate at the end of the 3rd iteration, and might need extra capacity for improving further.

Different from our previous experiences training wav2vec~2.0 models, we find that our \mmhubert{} models tend to be more stable during training. We do, however, reach training instability at the end of the 3rd iteration. For solving gradient explosion, similarly to wav2vec~2.0~(see discussion in \citet{PARCOLLET2024101622}), we find that the most effective solution is increasing floating point precision and restarting the optimizer from the previous learning rate.

\subsection{Pre-training hyper-parameters}
\label{appendix:pretraining:settings}

In this section, we cover the different hyper-parameters chosen for the training of \mmhubert{}, comparing intermediate checkpoints obtained using different settings. As a proxy evaluation of the quality of the speech representations obtained, we use the 1\,h phoneme recognition task from~\citet{9054548}. This setting allows us to investigate the emergence of multilingual phonemic content on our models through different iterations and training updates.

\paragraph{Proxy evaluation.} We train monolingual few-shot phoneme recognition systems using the 10 available languages: Spanish~(es), French~(fr), Italian~(it), Kyrgyz~(ky), Dutch~(nl), Russian~(ru), Swedish~(sv), Turkish~(tt), Tatar~(tt), Chinese~(zh-HK). We follow the standard data split, using 1\,h for training, validation and test. Our models are trained using the \texttt{Speechbrain} library~\cite{speechbrain}, with a standard CTC fine-tuning approach employing two different learning rates: $1e-5$ for the pre-trained SSL model, and $1e-4$ for the preceding 2 linear layers and output projection. We use a dropout of 0.6 and train for 100 epochs.\footnote{Recipe at: \url{https://github.com/utter-project/mHuBERT-147-scripts/tree/main/06_evaluation/CV_PR}.}

\begin{figure*}
\centering
\resizebox{\textwidth}{!}{
\includegraphics{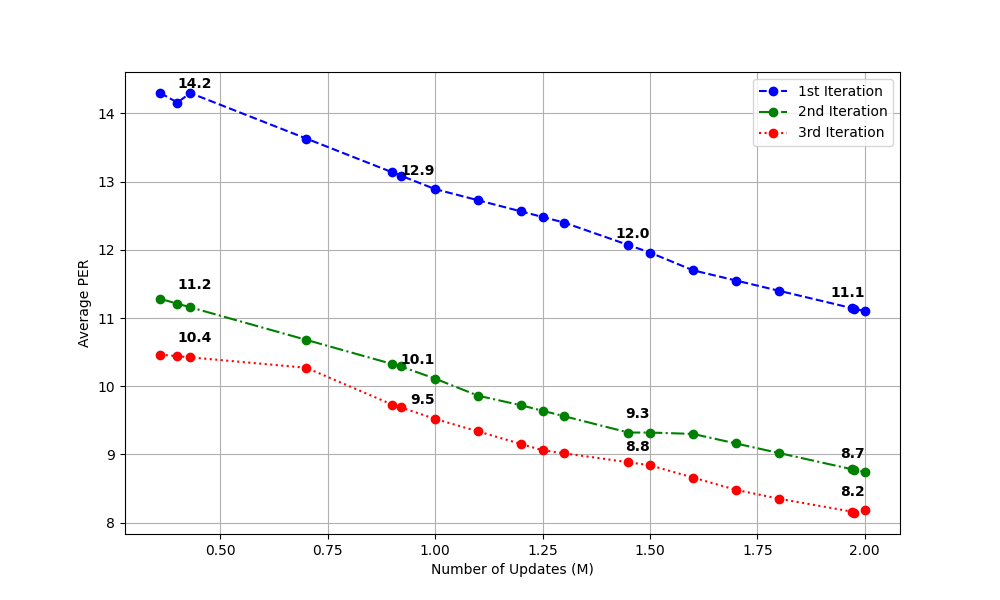}}
    \caption{Linear interpolation of average PER performance of \mmhubert{} across different iterations, and with different number of updates. The bold scores correspond to scores obtained at the following updates, from left to right: 400\,K, 1\,M, 1.5\,M and 2\,M.}
\label{fig:appendix:iterationvsupdates}
\end{figure*}
\begin{table*}[]
\centering
\begin{tabular}{lc|cccccccccc}
\toprule
        & \begin{tabular}[c]{@{}c@{}}\textbf{Average} \\ \textbf{PER ($\downarrow$)}\end{tabular} & \textbf{es}   & \textbf{fr}   & \textbf{it}   & \textbf{ky}   & \textbf{nl}   & \textbf{ru}   & \textbf{sv}   & \textbf{tr}   & \textbf{tt}  & \textbf{zh-HK} \\\midrule
\texttt{sklearn ($K=500$)} & 14.8& 10.8 & 13.7 & 15.0 & 11.5 & 15.8 & 16.6 & 19.2 & 13.2 & 9.5 & 22.3  \\
\texttt{faiss ($K=1000$)} & 14.6 & 10.7 & 13.6 & 14.8 & 10.9 & 15.5 & 16.4 & 19.3 & 13.2 & 9.4 & 22.5 \\\bottomrule
\toprule
\texttt{random crop} & 14.4 & 10.4 & 12.9 & 14.1 & 10.9 & 14.5 & 16.2 & 19.3 & 13.1 & 9.4 &  22.8 \\
\texttt{padding}   & 23.6 & 18.0 & 23.7 & 22.9 & 20.1 & 25.4 & 25.0 & 29.7 & 23.0 & 17.1 & 30.9 \\\bottomrule

\end{tabular}
\caption{PER scores for the proxy task. Models on top and bottom portions are not comparable due to different up-sampling settings and maximum number of updates~(top: 800\,K, bottom: 2\,M). \textbf{Top portion}: 1st iteration \mmhubert{} models~($\alpha=0.5$; $\beta=0.8$) trained for 800\,K updates using different clustering implementations: \texttt{sklearn} mini-batch k-means~($K=500$), or \texttt{faiss} IVF~($K=1000$). \textbf{Bottom portion}: 1st iteration \texttt{faiss}-based \mmhubert{} models~($\alpha=0.7$; $\beta=0.9$) trained for 400\,K updates using random crop or padding as batching approach.}
\label{appendix:kmeansfaissmultilingual}
\end{table*}

\subsubsection{Impact of number of updates and iterations}
\label{appendix:impact:updates}
We train our iterations for 2\,M updates, which is considerably larger than the standard approach (between 100-500\,K updates~\cite{hsu2021hubert,lee-etal-2022-textless}). In this section, we present evidence that pre-training longer is necessary for improved multilingual representation learning.

Figure~\ref{fig:appendix:iterationvsupdates} presents \mmhubert{} proxy performance across different iterations, and on varying number of updates.\footnote{Linear interpolation is used to build our plot as evaluation across updates did not always fetch checkpoints corresponding to the exact same number of updates. Nonetheless, we highlight in bold the points at which checkpoints were evaluated across all iterations.} Results are average PER scores computed across ten languages. We highlight performance at 400\,K, 1\,M, 1.5\,M and 2\,M.

Despite the loss curve looking stable after 400\,K~(Figure~\ref{fig:training:losscurves}), we observe that the models keep improving internal multilingual phonemic representation until the very end of the 2\,M updates. This is true for all iterations but the 3rd, where we achieve for the first time a best score that does not correspond to the last checkpoint: 8.1~PER at update 1.975\,M. We believe this is probably an artifact of the multilingual batching not favoring the specific subset of languages present in the proxy evaluation during the very last training batches. This can also be a result linked to model saturation.

Focusing on proxy results for the 1st and 2nd iterations, we observe that after 400\,K updates, the 2nd iteration model produces scores similar to that at the end of the 1st iteration. A longer training allows this model to yield an overall performance improvement of 21.6\% over the best score for the 1st iteration. 
Regarding the 3rd iteration, we observe that the proxy task improvement is not as significant compared to the 2nd iteration: the 3rd iteration model only surpasses the best 2nd iteration scores after 1.5\,M updates. This hints at the saturation of the multi-iteration HuBERT training approach at the 3rd step. We believe that more capacity~(i.e. going from base to large architecture) could allow the model to improve more substantially at this point.

Finally, we note that although this proxy task allows us to assess the model's potential during training, its scores are not completely correlated to the final model's downstream performance, mostly due to the limitation in covered languages~(only 10). For instance, although the 3rd iteration seems to improve only 0.5~PER over the second iteration, we observe significant improvement of ML-SUPERB scores when using the 3rd over the 2nd iteration model~(Section~\ref{sec:mlsuperb}).

\subsubsection{K-means versus faiss on multilingual settings}\label{sec:app:faissvskmeansmultilingual}

We train two 1st iteration \mmhubert{} models for 800\,K updates, varying the K-means implementation between the \texttt{sklearn}~($K=500$) and \texttt{faiss}~($K=1000$). We select $K=500$ for \texttt{sklearn} because, as mentioned in Section~\ref{section:3:faissvssklearn}, we were unable to train \texttt{sklearn} K-means models for our dataset using larger K values. Comparing models with different K values could result in a more favorable setting for \texttt{faiss}, but we believe this comparison can still be made using 1st iteration models. This is because at this step, clustering is performed using MFCCs, which are low-dimensional. Thus, the \textit{extra expressivity} that faiss-based clustering has at this iteration should not have a significant impact in the quality of the produced discretization.

The top portion of Table~\ref{appendix:kmeansfaissmultilingual} presents the results using our proxy evaluation setting for the two pre-trained 1st iteration models.\footnote{We highlight that these scores are not comparable to the ones in Section~\ref{appendix:impact:updates}, as for this experiment, the up-sampling is more aggressive ($\alpha=0.5$; $\beta=0.8$).} We once again observe that \texttt{faiss}-based \mmhubert{} training does not result in inferior speech representation modeling.

\subsubsection{Padding versus random crop}
We investigate the impact of random crop during multilingual training, compared to the more classical approach of padding. 
We train a first iteration padded \mmhubert{} model for 400\,K updates, and compare it to the model using random crop at the same training state~(same number of seen examples). The bottom portion of Table~\ref{appendix:kmeansfaissmultilingual} presents results for our proxy task. We observe that speech representation learning equipped with random crop batching mechanism outperforms quite drastically the padding approach at lower number of updates~(400\,K). We believe this is due to the higher example diversity for the random crop setting.

\subsubsection{Two-level language-source up-sampling}

The hyper-parameters for the two-level language-source up-sampling were chosen considering the language-source distribution in our data collection. The following are our observations regarding these hyper-parameters.

\paragraph{Language up-sampling.} Aggressive language up-sampling will result in a better multilingual balance in terms of language exposure, but it can result in poor utterance diversity during a given epoch. This issue comes from the fact that low-resource languages have a limited amount of examples, and thus aggressive up-sampling will allocate a large chunk of the batches for the same examples. While repeating examples in low-resource languages is the goal of up-sampling, we find that over-exposure to the same limited number of examples will result in overall multilingual speech representation degradation. For this reason, during our hyper-parameter search, we moved from $\alpha=0.5$~(41\% of a given epoch allocated to repeated examples) to $\alpha=0.7$~(20\% of a given epoch allocated to repeated examples).

\paragraph{Source up-sampling.} We find that this parameter is more important for high-resource languages, as low-resource languages are often single-sourced in our multilingual dataset. As we filtered overly large datasets for high-resource languages prior to training~(Section~\ref{sec:data}), we employ a mild source up-sampling.

\subsection{Open-source links}\label{appendix:section:opensource}

The \mmhubert{} models are made available under the CC-BY-NC-SA-4.0 license. Our model release include the \texttt{faiss} trained IVFs, the \texttt{fairseq} checkpoint and the corresponding \texttt{HuggingFace} checkpoint.

\begin{itemize}
    \item \mmhubert{} (\texttt{-3rd}): \url{https://huggingface.co/utter-project/mHuBERT-147}
    \item \mmhubert{}\texttt{-2nd}: \url{https://huggingface.co/utter-project/mHuBERT-147-base-2nd-iter}
    \item \mmhubert{}\texttt{-1st}: \url{https://huggingface.co/utter-project/mHuBERT-147-base-1st-iter}
    \item Manifest files: \url{https://huggingface.co/utter-project/mHuBERT-147-base-3rd-iter/tree/main/manifest}
    \item Training scripts (\texttt{fairseq} fork): \url{https://github.com/utter-project/fairseq}
    \item Pre-processing and evaluation scripts (including \texttt{faiss} scripts): \url{https://github.com/utter-project/mHuBERT-147-scripts}
\end{itemize}

\paragraph{Intermediate checkpoints.} Intermediate checkpoints for the three iterations are made available under the \textbf{CC-BY-NC-SA-4.0} license via a protected link. 
\begin{itemize}
    \item \textbf{user:} user
    \item \textbf{password:} the license mentioned above
    \item \textbf{URL:} \url{https://download.europe.naverlabs.com/mhubert147/}
\end{itemize}
\clearpage
\section{Downstream experiments}
\label{app.downstream}

\subsection{ML-SUPERB experiments}\label{appendix:experiments:mlsuperb}

For evaluation on ML-SUPERB, we follow the recipe provided in the \texttt{ESPnet} library~\cite{watanabe2018espnet}. We use their \texttt{Hugging Face} interface for fine-tuning our \mmhubert{} model.\footnote{More information available at: \url{https://github.com/espnet/espnet/tree/master/egs2/ml_superb/asr1}} We do implement an ML-SUPERB score calculator, since we could not find one provided by the organizers.\footnote{Available at: \url{https://github.com/utter-project/mHuBERT-147-scripts/blob/main/06_evaluation/mlsuperb/compute_superb_score.py}}

\paragraph{Learning Rate.}
Following the official ML-SUPERB guidelines, we only optimize the learning rate parameter~(best of $1e-3;1e-4;1e-5$). Table~\ref{tab:appendix:mlsuperb:learningrates} presents the list of learning rates selected by evaluating models on the validation set. The scores presented in Table~\ref{tab:mlsuperb} correspond to models trained with the corresponding learning rate.
\\We highlight that our experimentation showed that it is quite important to evaluate models with multiple learning rates. We show test LID results in Table~\ref{tab:appendix:mlsuperb:lid} for \mmhubert{}\texttt{-3rd} trained using different learning rates. By varying the learning rate, we go from mediocre LID results to SOTA scores for this ML-SUPERB task.

\paragraph{Monolingual ASR Task.}
Table~\ref{tab:appendix:mlsuperb:monolingual} present test CER results per language for \mmhubert{}\texttt{-3rd}, in both 10\,min and 1\,h settings. For each language, we train three models, varying the learning rate. We use the development set to select the final model for our submission~(see Table~\ref{tab:appendix:mlsuperb:learningrates}). During training, we observed many instabilities for this particular setting, with models sometimes failing to converge. In these cases, we find that restarting training with a new random seed fixed the problem. We highlight that seeds vary across languages.

\begin{table}
\centering
\resizebox{\columnwidth}{!}{
\begin{tabular}{lcccc}\toprule
\multicolumn{1}{c}{\textbf{}}   & \textbf{\begin{tabular}[c]{@{}c@{}}Monolingual \\ ASR\end{tabular}}                                  & \textbf{\begin{tabular}[c]{@{}c@{}}Multilingual \\ ASR\end{tabular}} & \textbf{LID}              & \textbf{\begin{tabular}[c]{@{}c@{}}Multilingual \\ ASR+LID\end{tabular}} \\\midrule
\multirow{2}{*}{\textbf{10\,min}} & \begin{tabular}[c]{@{}c@{}}1e-3\\ (GROUP A)\end{tabular} & 1e-3 & 1e-4 & 1e-3\\
& \begin{tabular}[c]{@{}c@{}}1e-4\\ (GROUP B)\end{tabular}  & & & \\\midrule
\textbf{1\,h}  & 1e-3  & 1e-3 & 1e-4   & 1e-4    \\\bottomrule               
\end{tabular}}
\caption{Selected learning rates for \mmhubert{}\texttt{-3rd} using validation scores for the corresponding tasks. For monolingual ASR, \textbf{GROUP A} corresponds to: eng1, eng2, eng3, deu1, rus, swa, jpn, cmn, and \textbf{GROUP B} corresponds to: fra1, fra2, deu2, swe, xty.}
\label{tab:appendix:mlsuperb:learningrates}
\end{table}
\begin{table}
\centering
\begin{tabular}{lcc}\toprule
&               & \textbf{ACC ($\uparrow$)}   \\\midrule
\multirow{3}{*}{\textbf{10\,min}} & 1e-3 & 49.68          \\
& 1e-4 & \textbf{85.33} \\
& 1e-5 & 69.56          \\\midrule
\multirow{3}{*}{\textbf{1\,h}}    & 1e-3 & 51.06          \\
& 1e-4 & \textbf{90.95} \\
& 1e-5 & 81.35     \\\bottomrule    
\end{tabular}
\caption{ML-SUPERB 10\,min and 1\,h LID ACC test scores for \mmhubert{}\texttt{-3rd} by varying the learning rate for training.}
\label{tab:appendix:mlsuperb:lid}
\end{table}
\begin{table*}
\centering
\resizebox{\textwidth}{!}{\begin{tabular}{cc|ccc|cc|cc|cccccc}\toprule
               & \textbf{\begin{tabular}[c]{@{}c@{}}Final\\ Score\end{tabular}} & \textbf{eng1} & \textbf{eng2} & \textbf{eng3} & \textbf{fra1} & \textbf{fra2} & \textbf{deu1} & \textbf{deu2} & \textbf{rus} & \textbf{swa} & \textbf{swe} & \textbf{jpn} & \textbf{cmn} & \textbf{xty} \\\midrule
\textbf{10min} & 34.2                                                           & 30.9          & 42.3          & 37.0          & 43.2          & 51.0          & 25.2          & 36.9          & 26.4         & 24.7         & 29.9         & 13.9         & 34.7         & 63.0         \\
\textbf{1h}    & 26.3                                                           & 22.6          & 33.6          & 27.3          & 29.5          & 36.1          & 20.2         & 24.3         & 20.5         & 19.8         & 21.6         & 10.4         & 23.8         & 57.9      \\\bottomrule
\end{tabular}}
\caption{Detailed \mmhubert{}\texttt{-3rd} test CER ($\downarrow$) results for  the languages inside the ML-SUPERB monolingual track. The final score is obtained by averaging all runs belonging to the same language, and then averaging this score across all different languages.}
\label{tab:appendix:mlsuperb:monolingual}
\end{table*}

\subsection{FLEURS-102 experiments}\label{appendix:experiments:fleurs102}

Table~\ref{appendix:tab:fleurs102} present average CER results for all languages in FLEURS-102. 

\paragraph{Failed ASR models.} We find that MMS and XLS-R models fail to converge to useful translation~(CER$>90$) for 16 languages (ara, ceb, ckb, cym, fas, ful, heb, hrv, hye, mri, nya, oci, snd, som, tam, tel). For \mmhubert{}, this only happens for half of these languages~(cym, fas, hye, nya, oci, snd, som, tel). There is no instance where \mmhubert{} fails to converge but other models successfully converge.

\paragraph{Seen versus unseen languages.} We use the model closest to ours in language coverage~(XLS-R) to compare model performance for seen and unseen languages. Table~\ref{appendix:table:xlsrmhubert} presents the results. We observe that while being much more compact, \mmhubert{} presents overall better performance in both seen and unseen languages.

\begin{table*}
\centering
\scriptsize
\resizebox{\textwidth}{!}{\begin{tabular}{lcccccccccccc}
\toprule
       & afr              & amh              & ara              & asm              & ast              & aze              & bel              & ben              & bos              & bul              & cat              & ceb              \\ \midrule
MMS-1B & \textbf{9.5}$\pm$\tiny{0.02}  & \textbf{7.9}$\pm$\tiny{0.13}  & 99.1$\pm$\tiny{0.0}   & \textbf{9.8}$\pm$\tiny{0.02}  & \textbf{5.7}$\pm$\tiny{0.13}  & \textbf{6.0}$\pm$\tiny{0.01}  & \textbf{4.4}$\pm$\tiny{0.1}   & \textbf{7.0}$\pm$\tiny{0.07}  & \textbf{5.0}$\pm$\tiny{0.03}  & \textbf{4.2}$\pm$\tiny{0.02}  & \textbf{4.5}$\pm$\tiny{0.01}  & 86.8$\pm$\tiny{6.43} \\ 
MMS-300M & 12.5$\pm$\tiny{0.09} & 9.4$\pm$\tiny{0.13}  & 98.7$\pm$\tiny{0.37} & 12.4$\pm$\tiny{0.08} & 7.0$\pm$\tiny{0.01}  & 7.8$\pm$\tiny{0.03}  & 5.8$\pm$\tiny{0.0}   & 8.0$\pm$\tiny{0.1}   & 6.5$\pm$\tiny{0.08}  & 5.6$\pm$\tiny{0.05}  & 6.1$\pm$\tiny{0.08}  & 99.3$\pm$\tiny{0.0}   \\ 
XLS-R  & 12.0$\pm$\tiny{0.73} & 10.6$\pm$\tiny{0.31} & 99.1$\pm$\tiny{0.0}   & 12.6$\pm$\tiny{0.07} & 6.4$\pm$\tiny{0.02}  & 8.0$\pm$\tiny{0.15}  & 5.6$\pm$\tiny{0.04}  & 8.4$\pm$\tiny{0.0}   & 5.7$\pm$\tiny{0.09}  & 4.8$\pm$\tiny{0.18}  & 5.2$\pm$\tiny{0.04}  & 99.3$\pm$\tiny{0.0}   \\ 
\mmhubert{} & 15.3$\pm$\tiny{0.74} & 12.1$\pm$\tiny{0.21} & \textbf{9.8}$\pm$\tiny{0.16}  & 15.1$\pm$\tiny{0.18} & 10.2$\pm$\tiny{0.24} & 11.3$\pm$\tiny{0.33} & 8.0$\pm$\tiny{0.11}  & 9.6$\pm$\tiny{0.13}  & 9.3$\pm$\tiny{0.13}  & 8.5$\pm$\tiny{0.05}  & 8.4$\pm$\tiny{0.02}  & \textbf{57.0}$\pm$\tiny{43.03} \\ \bottomrule
\end{tabular}}
\resizebox{\textwidth}{!}{\begin{tabular}{lcccccccccccc}
\toprule
       & ces              & ckb              & cmn              & cym              & dan              & deu              & ell              & eng              & est              & fas              & fil              & fin              \\ \midrule
MMS-1B & \textbf{4.6}$\pm$\tiny{0.07}  & 97.4$\pm$\tiny{1.82}  & \textbf{17.9}$\pm$\tiny{0.58}  & \textbf{99.3}$\pm$\tiny{0.0}  & \textbf{9.1}$\pm$\tiny{0.11}  & \textbf{3.9}$\pm$\tiny{0.02}  & \textbf{6.1}$\pm$\tiny{0.35}  & \textbf{6.4}$\pm$\tiny{0.26}  & \textbf{3.5}$\pm$\tiny{0.04}  & 99.6$\pm$\tiny{0.0}  & \textbf{4.4}$\pm$\tiny{0.11}  & \textbf{2.9}$\pm$\tiny{0.06} \\ 
MMS-300M & 6.8$\pm$\tiny{0.04}  & 98.0$\pm$\tiny{1.24}  & 30.1$\pm$\tiny{0.07}  & \textbf{99.3}$\pm$\tiny{0.0}  & 12.4$\pm$\tiny{0.02}  & 6.0$\pm$\tiny{0.1}  & 8.3$\pm$\tiny{0.2}  & 8.8$\pm$\tiny{0.4}  & 4.4$\pm$\tiny{0.15}  & 99.6$\pm$\tiny{0.0}  & 5.4$\pm$\tiny{0.14}  & 3.7$\pm$\tiny{0.02} \\ 
XLS-R-300M & 5.7$\pm$\tiny{0.08}  & 94.9$\pm$\tiny{4.32}  & 27.9$\pm$\tiny{1.63}  & \textbf{99.3}$\pm$\tiny{0.0}  & 11.1$\pm$\tiny{0.22}  & 5.0$\pm$\tiny{0.04}  & 7.5$\pm$\tiny{0.24}  & 7.8$\pm$\tiny{0.11}  & 3.9$\pm$\tiny{0.16}  & \textbf{96.0}$\pm$\tiny{3.59}  & 5.4$\pm$\tiny{0.13}  & 3.3$\pm$\tiny{0.04} \\ 
\mmhubert{} & 11.1$\pm$\tiny{0.1}  & \textbf{58.5}$\pm$\tiny{41.52}  & 30.7$\pm$\tiny{1.15}  & \textbf{99.3}$\pm$\tiny{0.0}  & 17.6$\pm$\tiny{0.09}  & 9.6$\pm$\tiny{0.14}  & 13.5$\pm$\tiny{0.04}  & 11.0$\pm$\tiny{0.14}  & 6.4$\pm$\tiny{0.04}  & 100.0$\pm$\tiny{0.0}  & 7.2$\pm$\tiny{0.2}  & 5.4$\pm$\tiny{0.01} \\ \bottomrule
\end{tabular}}
\resizebox{\textwidth}{!}{\begin{tabular}{l *{12}{c}}
        \toprule
        & fra & ful & gle & glg & guj & hau & heb & hin & hrv & hun & hye & ibo \\
        \midrule
        MMS-1B     & \textbf{7.0}$\pm$\tiny{0.1}    & 98.4$\pm$\tiny{0.86}   & \textbf{24.3}$\pm$\tiny{0.1}    & \textbf{3.8}$\pm$\tiny{0.09}   & \textbf{6.7}$\pm$\tiny{0.24}   & \textbf{8.2}$\pm$\tiny{0.06}   & 99.5$\pm$\tiny{0.12}  & \textbf{7.2}$\pm$\tiny{0.08}   & 99.4$\pm$\tiny{0.0}   & \textbf{5.8}$\pm$\tiny{0.02}   & 99.3$\pm$\tiny{0.0}   & \textbf{13.1}$\pm$\tiny{0.02}  \\
        MMS-300M   & 10.5$\pm$\tiny{0.16}  & 103.8$\pm$\tiny{4.62}  & 29.2$\pm$\tiny{0.08}   & 5.1$\pm$\tiny{0.0}     & 8.0$\pm$\tiny{0.02}     & 9.2$\pm$\tiny{0.01}     & 109.4$\pm$\tiny{10.43}  & 9.3$\pm$\tiny{0.06}     & 99.4$\pm$\tiny{0.0}     & 9.2$\pm$\tiny{0.09}     & 98.6$\pm$\tiny{0.67}     & 14.7$\pm$\tiny{0.11}     \\
        XLS-R-300M      & 9.0$\pm$\tiny{0.21}    & 94.8$\pm$\tiny{3.23}    & 29.1$\pm$\tiny{0.34}   & 4.6$\pm$\tiny{0.03}     & 8.5$\pm$\tiny{0.12}     & 10.1$\pm$\tiny{0.11}    & 107.5$\pm$\tiny{0.6}    & 10.3$\pm$\tiny{0.15}    & 95.8$\pm$\tiny{3.62}    & 8.7$\pm$\tiny{0.22}     & \textbf{96.2}$\pm$\tiny{3.13}     & 15.0$\pm$\tiny{0.2}      \\
        \mmhubert{} & 16.1$\pm$\tiny{0.08}  & \textbf{19.1}$\pm$\tiny{0.03}   & 32.7$\pm$\tiny{0.08}   & 8.3$\pm$\tiny{0.05}    & 10.4$\pm$\tiny{0.16}    & 12.0$\pm$\tiny{0.19}    & \textbf{22.5}$\pm$\tiny{0.02}    & 12.5$\pm$\tiny{0.25}    & \textbf{8.5}$\pm$\tiny{0.05}     & 14.1$\pm$\tiny{0.23}    & 100.0$\pm$\tiny{0.0}    & 17.1$\pm$\tiny{0.03}     \\
        \bottomrule
    \end{tabular}}
\resizebox{\textwidth}{!}{\begin{tabular}{l *{12}{c}}
    \toprule
    & ind & isl & ita & jav & jpn & kam & kan & kat & kaz & kea & khm & kir \\
    \midrule
    MMS-1B     & \textbf{3.9}$\pm$\tiny{0.09}  & \textbf{9.4}$\pm$\tiny{0.01}  & \textbf{2.1}$\pm$\tiny{0.01}  & \textbf{5.4}$\pm$\tiny{0.23}  & \textbf{22.2}$\pm$\tiny{0.26}  & \textbf{12.5}$\pm$\tiny{0.06}  & \textbf{5.8}$\pm$\tiny{0.01}  & \textbf{5.6}$\pm$\tiny{0.06}  & \textbf{3.3}$\pm$\tiny{0.08}  & \textbf{4.9}$\pm$\tiny{0.02}  & \textbf{16.4}$\pm$\tiny{0.07}  & \textbf{4.6}$\pm$\tiny{0.14} \\
    MMS-300M   & 4.8$\pm$\tiny{0.1}  & 15.5$\pm$\tiny{0.07}  & 2.8$\pm$\tiny{0.04}  & 6.8$\pm$\tiny{0.15}  & 33.3$\pm$\tiny{0.29}  & 13.8$\pm$\tiny{0.39}  & 7.3$\pm$\tiny{0.12}  & 8.0$\pm$\tiny{0.08}  & 4.1$\pm$\tiny{0.01}  & 5.8$\pm$\tiny{0.03}  & 22.0$\pm$\tiny{0.14}  & 5.8$\pm$\tiny{0.15} \\
    XLS-R-300M     & 4.8$\pm$\tiny{0.21}  & 16.0$\pm$\tiny{0.25}  & 2.4$\pm$\tiny{0.01}  & 7.1$\pm$\tiny{0.19}  & 34.9$\pm$\tiny{0.17}  & 15.0$\pm$\tiny{0.15}  & 7.8$\pm$\tiny{0.04}  & 7.9$\pm$\tiny{0.03}  & 4.0$\pm$\tiny{0.16}  & 5.3$\pm$\tiny{0.03}  & 22.5$\pm$\tiny{0.85}  & 6.1$\pm$\tiny{0.05} \\
    \mmhubert{} & 6.8$\pm$\tiny{0.15}  & 14.9$\pm$\tiny{0.27}  & 4.8$\pm$\tiny{0.06}  & 9.6$\pm$\tiny{0.09}  & 38.4$\pm$\tiny{0.73}  & 16.3$\pm$\tiny{0.08}  & 8.5$\pm$\tiny{0.12}  & 10.3$\pm$\tiny{0.19}  & 5.6$\pm$\tiny{0.17}  & 7.8$\pm$\tiny{0.01}  & 24.5$\pm$\tiny{0.11}  & 7.5$\pm$\tiny{0.08} \\
    \bottomrule
\end{tabular}}
\resizebox{\textwidth}{!}{\begin{tabular}{l *{12}{c}}
    \toprule
    & kor & lao & lav & lin & lit & ltz & lug & luo & mal & mar & mkd & mlt \\
    \midrule
    MMS-1B     & \textbf{15.0}$\pm$\tiny{0.14}  & \textbf{24.9}$\pm$\tiny{0.67}  & \textbf{3.4}$\pm$\tiny{0.13}  & \textbf{5.1}$\pm$\tiny{0.09}  & \textbf{5.0}$\pm$\tiny{0.01}  & \textbf{8.9}$\pm$\tiny{0.22}  & \textbf{9.5}$\pm$\tiny{0.14}  & \textbf{6.3}$\pm$\tiny{0.05}  & \textbf{5.3}$\pm$\tiny{0.19}  & \textbf{8.7}$\pm$\tiny{0.14}  & \textbf{2.6}$\pm$\tiny{0.01}  & \textbf{4.9}$\pm$\tiny{0.06} \\
    MMS-300M   & 21.7$\pm$\tiny{0.5}  & 29.5$\pm$\tiny{0.22}  & 4.7$\pm$\tiny{0.09}  & 5.9$\pm$\tiny{0.12}  & 7.1$\pm$\tiny{0.03}  & 11.1$\pm$\tiny{0.11}  & 10.2$\pm$\tiny{0.01}  & 7.3$\pm$\tiny{0.07}  & 6.3$\pm$\tiny{0.05}  & 11.2$\pm$\tiny{0.07}  & 3.4$\pm$\tiny{0.08}  & 6.2$\pm$\tiny{0.04} \\
    XLS-R-300M & 24.6$\pm$\tiny{1.37}  & 30.4$\pm$\tiny{0.15}  & 4.1$\pm$\tiny{0.08}  & 6.0$\pm$\tiny{0.13}  & 6.1$\pm$\tiny{0.01}  & 10.7$\pm$\tiny{0.01}  & 10.6$\pm$\tiny{0.02}  & 8.1$\pm$\tiny{0.12}  & 6.6$\pm$\tiny{0.14}  & 12.2$\pm$\tiny{0.02}  & 3.2$\pm$\tiny{0.02}  & 5.4$\pm$\tiny{0.01} \\
    \mmhubert{} & 20.7$\pm$\tiny{0.22}  & 33.5$\pm$\tiny{0.17}  & 7.6$\pm$\tiny{0.1}  & 7.6$\pm$\tiny{0.06}  & 10.2$\pm$\tiny{0.07}  & 14.9$\pm$\tiny{0.03}  & 11.8$\pm$\tiny{0.15}  & 9.3$\pm$\tiny{0.22}  & 7.6$\pm$\tiny{0.05}  & 14.1$\pm$\tiny{0.15}  & 5.1$\pm$\tiny{0.18}  & 8.7$\pm$\tiny{0.13} \\
    \bottomrule
\end{tabular}
}
\resizebox{\textwidth}{!}{\begin{tabular}{l *{12}{c}}
    \toprule
    & mon & mri & msa & mya & nep & nld & nob & nso & nya & oci & ori & orm \\
    \midrule
    MMS-1B     & \textbf{8.6}$\pm$\tiny{0.23}  & 99.3$\pm$\tiny{0.0}  & \textbf{4.6}$\pm$\tiny{0.02}  & \textbf{15.7}$\pm$\tiny{0.48}  & \textbf{10.0}$\pm$\tiny{0.25}  & \textbf{5.3}$\pm$\tiny{0.12}  & \textbf{5.2}$\pm$\tiny{0.04}  & \textbf{8.0}$\pm$\tiny{0.13}  & \textbf{99.3}$\pm$\tiny{0.0}  & \textbf{96.3}$\pm$\tiny{2.79}  & \textbf{11.3}$\pm$\tiny{0.12}  & \textbf{16.8}$\pm$\tiny{0.37} \\
    MMS-300M   & 11.2$\pm$\tiny{0.04}  & 99.7$\pm$\tiny{0.35}  & 6.7$\pm$\tiny{0.05}  & 20.6$\pm$\tiny{0.2}  & 12.4$\pm$\tiny{0.34}  & 8.0$\pm$\tiny{0.16}  & 6.5$\pm$\tiny{0.05}  & 9.7$\pm$\tiny{0.11}  & \textbf{99.3}$\pm$\tiny{0.0}  & \textbf{99.4}$\pm$\tiny{0.0}  & 16.2$\pm$\tiny{0.09}  & 19.6$\pm$\tiny{0.17} \\
    XLS-R-300M & 11.0$\pm$\tiny{0.32}  & 96.7$\pm$\tiny{0.24}  & 7.0$\pm$\tiny{0.09}  & 20.2$\pm$\tiny{0.11}  & 13.2$\pm$\tiny{0.47}  & 6.6$\pm$\tiny{0.14}  & 6.0$\pm$\tiny{0.13}  & 10.3$\pm$\tiny{0.21}  & \textbf{99.3}$\pm$\tiny{0.0}  & \textbf{99.4}$\pm$\tiny{0.0}  & 18.4$\pm$\tiny{0.05}  & 20.0$\pm$\tiny{0.47} \\
    \mmhubert{} & 14.0$\pm$\tiny{0.43}  & \textbf{11.5}$\pm$\tiny{0.01}  & 9.9$\pm$\tiny{0.16}  & 22.3$\pm$\tiny{0.12}  & 15.4$\pm$\tiny{0.19}  & 12.2$\pm$\tiny{0.26}  & 9.5$\pm$\tiny{0.16}  & 12.5$\pm$\tiny{0.07}  & \textbf{99.3}$\pm$\tiny{0.0}  & \textbf{99.4}$\pm$\tiny{0.0}  & 19.6$\pm$\tiny{0.1}  & 21.1$\pm$\tiny{0.3} \\
    \bottomrule
\end{tabular}
}
\resizebox{\textwidth}{!}{\begin{tabular}{l *{12}{c}}
    \toprule
    & pan & pol & por & pus & ron & rus & slk & slv & sna & snd & som & spa \\
    \midrule
    MMS-1B     & \textbf{9.1}$\pm$\tiny{0.03}  & \textbf{4.2}$\pm$\tiny{0.01}  & \textbf{4.4}$\pm$\tiny{0.08}  & \textbf{18.3}$\pm$\tiny{0.12}  & \textbf{5.0}$\pm$\tiny{0.03}  & \textbf{4.8}$\pm$\tiny{0.01}  & \textbf{3.5}$\pm$\tiny{0.07}  & \textbf{6.2}$\pm$\tiny{0.0}  & \textbf{5.0}$\pm$\tiny{0.08}  & 99.1$\pm$\tiny{0.0}  & 99.3$\pm$\tiny{0.0}  & \textbf{2.9}$\pm$\tiny{0.08} \\
    MMS-300M   & 11.4$\pm$\tiny{0.14}  & 6.0$\pm$\tiny{0.01}  & 5.8$\pm$\tiny{0.02}  & 21.0$\pm$\tiny{0.23}  & 7.2$\pm$\tiny{0.02}  & 6.5$\pm$\tiny{0.07}  & 4.7$\pm$\tiny{0.1}  & 8.2$\pm$\tiny{0.19}  & 5.9$\pm$\tiny{0.02}  & \textbf{96.1}$\pm$\tiny{3.05}  & 97.7$\pm$\tiny{1.69}  & 3.8$\pm$\tiny{0.03} \\
    XLS-R-300M & 12.4$\pm$\tiny{0.15}  & 5.0$\pm$\tiny{0.05}  & 4.9$\pm$\tiny{0.04}  & 21.5$\pm$\tiny{0.51}  & 6.3$\pm$\tiny{0.05}  & 6.0$\pm$\tiny{0.15}  & 4.0$\pm$\tiny{0.07}  & 7.2$\pm$\tiny{0.11}  & 6.5$\pm$\tiny{0.03}  & 99.1$\pm$\tiny{0.0}  & \textbf{96.7}$\pm$\tiny{2.66}  & 3.3$\pm$\tiny{0.05} \\
    \mmhubert{} & 14.5$\pm$\tiny{0.07}  & 10.2$\pm$\tiny{0.18}  & 9.0$\pm$\tiny{0.26}  & 24.0$\pm$\tiny{0.29}  & 10.9$\pm$\tiny{0.0}  & 9.2$\pm$\tiny{0.08}  & 7.6$\pm$\tiny{0.07}  & 11.1$\pm$\tiny{0.02}  & 8.5$\pm$\tiny{0.19}  & 99.1$\pm$\tiny{0.0}  & 100.0$\pm$\tiny{0.0}  & 6.1$\pm$\tiny{0.08} \\
    \bottomrule
\end{tabular}
}
\resizebox{\textwidth}{!}{\begin{tabular}{l *{12}{c}}
    \toprule
    & srp & swa & swe & tam & tel & tgk & tha & tur & ukr & umb & urd & uzb \\
    \midrule
    MMS-1B     & \textbf{14.2}$\pm$\tiny{0.11}  & \textbf{4.2}$\pm$\tiny{0.05}  & \textbf{7.0}$\pm$\tiny{0.23}  & 99.3$\pm$\tiny{0.0}  & \textbf{99.3}$\pm$\tiny{0.0}  & \textbf{5.1}$\pm$\tiny{0.04}  & \textbf{12.6}$\pm$\tiny{0.11}  & \textbf{4.3}$\pm$\tiny{0.16}  & \textbf{4.9}$\pm$\tiny{0.01}  & \textbf{16.6}$\pm$\tiny{0.27}  & \textbf{8.8}$\pm$\tiny{0.11}  & \textbf{7.7}$\pm$\tiny{0.04} \\
    MMS-300M   & 16.9$\pm$\tiny{0.17}  & 5.5$\pm$\tiny{0.06}  & 10.0$\pm$\tiny{0.06}  & 98.0$\pm$\tiny{1.32}  & \textbf{99.3}$\pm$\tiny{0.0}  & 5.9$\pm$\tiny{0.05}  & 16.1$\pm$\tiny{0.17}  & 5.8$\pm$\tiny{0.09}  & 6.8$\pm$\tiny{0.0}  & 18.9$\pm$\tiny{0.41}  & 10.9$\pm$\tiny{0.05}  & 10.3$\pm$\tiny{0.0} \\
    XLS-R-300M & 16.5$\pm$\tiny{0.66}  & 5.6$\pm$\tiny{0.1}  & 9.1$\pm$\tiny{0.01}  & 92.9$\pm$\tiny{6.42}  & \textbf{99.3}$\pm$\tiny{0.0}  & 6.2$\pm$\tiny{0.18}  & 16.0$\pm$\tiny{0.19}  & 5.8$\pm$\tiny{0.09}  & 6.1$\pm$\tiny{0.08}  & 20.0$\pm$\tiny{0.64}  & 13.6$\pm$\tiny{0.15}  & 10.7$\pm$\tiny{0.12} \\
    \mmhubert{} & 18.8$\pm$\tiny{0.15}  & 6.9$\pm$\tiny{0.04}  & 15.2$\pm$\tiny{0.11}  & \textbf{16.3}$\pm$\tiny{0.22}  & \textbf{99.3}$\pm$\tiny{0.0}  & 7.0$\pm$\tiny{0.08}  & 17.7$\pm$\tiny{0.12}  & 7.4$\pm$\tiny{0.07}  & 9.6$\pm$\tiny{0.0}  & 23.4$\pm$\tiny{0.12}  & 15.1$\pm$\tiny{0.36}  & 13.8$\pm$\tiny{0.27} \\
    \bottomrule
\end{tabular}}
\begin{tabular}{l *{6}{c}}
    \toprule
    & vie & wol & xho & yor & yue & zul \\
    \midrule
    MMS-1B     & \textbf{10.9}$\pm$\tiny{0.11}  & \textbf{14.7}$\pm$\tiny{0.1}  & \textbf{6.7}$\pm$\tiny{0.09}  & \textbf{18.2}$\pm$\tiny{0.22}  & \textbf{16.1}$\pm$\tiny{0.01}  & \textbf{6.6}$\pm$\tiny{0.03} \\
    MMS-300M   & 14.0$\pm$\tiny{0.36}  & 16.1$\pm$\tiny{0.29}  & 7.6$\pm$\tiny{0.02}  & 21.1$\pm$\tiny{0.17}  & 58.1$\pm$\tiny{19.39}  & 8.3$\pm$\tiny{0.15} \\
    XLS-R-300M & 15.6$\pm$\tiny{0.31}  & 16.2$\pm$\tiny{0.4}  & 8.0$\pm$\tiny{0.07}  & 22.1$\pm$\tiny{0.5}  & 46.3$\pm$\tiny{5.03}  & 8.5$\pm$\tiny{0.0} \\
    \mmhubert{} & 17.3$\pm$\tiny{0.14}  & 17.8$\pm$\tiny{0.01}  & 9.8$\pm$\tiny{0.05}  & 23.9$\pm$\tiny{0.08}  & 36.1$\pm$\tiny{1.55}  & 11.1$\pm$\tiny{0.01} \\
    \bottomrule
\end{tabular}
\caption{FLEURS-102 CER ($\downarrow$) average results per language with standard deviations. Best scores in \textbf{bold}.}
\label{appendix:tab:fleurs102}
\end{table*}

\begin{table}
\centering
\begin{tabular}{lc|cc}\toprule
            & \textbf{\begin{tabular}[c]{@{}c@{}}General \\ Avg (102)\end{tabular}} & \textbf{\begin{tabular}[c]{@{}c@{}}Seen \\ (87)\end{tabular}} & \textbf{\begin{tabular}[c]{@{}c@{}}Unseen \\ (15)\end{tabular}} \\\midrule
XLS-R-300M  & 24.5                                                                  & 23.8                                                          & 28.3                                                            \\
\mmhubert{} & \textbf{21.1}                                                         & \textbf{20.8}                                                 & \textbf{22.6} \\\bottomrule                                                  
\end{tabular}
\caption{FLEURS-102 CER ($\downarrow$) average scores for seen and unseen languages.}
\label{appendix:table:xlsrmhubert}
\end{table}

\subsection{Monolingual versus multilingual HuBERT}\label{appendix:experiments:monolingual}

In this section we compare the English speech representation from our best \mmhubert{} model against a model of same capacity, but trained on English only data~\cite{hsu2021hubert}.\footnote{Available at: \url{https://huggingface.co/facebook/hubert-base-ls960}} With these experiments, we want to illustrate that \mmhubert{} is not only competitive against larger multilingual speech representation models~(See Sections~\ref{sec:mlsuperb} and \ref{sec:fleurs102}), but it is also competitive to monolingual solutions that do not share their parameters across languages.

\paragraph{Setup.} We train ASR models for open datasets from the ESB benchmark~\cite{gandhi2022esb}. We use the data selection method from \citet{10446931} to downsample training data to 50\,h for faster training with marginal performance impact. We also adopt their experimental setup, which we now detail.
\\We train English ASR models using either \mmhubert{} or the original HuBERT-base~(\texttt{hubert-base-ls960}) as speech encoder, and follow the \texttt{SpeechBrain} library~\cite{speechbrain}'s recipe for ASR -- detailed in Appendix Section~\ref{appendix:pretraining:settings}~(proxy evaluation). In these experiments too, we train ASR models using two distinct optimizers, one for the upstream model, and another for the three subsequent neural networks that precede the CTC output projection layer~($lr = 0.9$). Table~\ref{tab:appendix:mono:hyper} presents hyper-parameters we use per dataset.

\paragraph{Results.} Table~\ref{tab:appendix:monolingual:multilingual} presents WER results for both models across all datasets. The average WER for the ASR models using HuBERT-base and \mmhubert{} is respectively 23.1 and 23.6. Unsurprisingly, we find that fine-tuning a pre-trained model trained solely for the target language~(HuBERT-base) yields better results for almost all datasets. Nonetheless, we observe that \mmhubert{}, despite being trained on multilingual settings and having to share its parameters across many languages, still performs competitively with the English model. This model is on average only 0.5 WER points worse than HuBERT-base. We believe that the competitiveness of \mmhubert{} compared to this monolingual model highlights its potential as a compact yet powerful backbone for multilingual speech applications.

\begin{table}
\centering
\resizebox{\columnwidth}{!}{
\begin{tabular}{cccccc}\toprule
            & \textbf{warm-up} & \textbf{learning rate} & \textbf{dropout} & \textbf{epochs} \\ 
            & \textbf{steps} & \textbf{upstream} & & \\\midrule
\textbf{AMI}~\cite{carletta2007unleashing,renals2007recognition}        & 200     & 5e-5  & 0.3  & 30  \\ 
\textbf{CV}~\cite{commonvoice:2020}         & 100     & 5e-5  & 0    & 30  \\ 
\textbf{Earnings-22}~\cite{del2022earnings} & 200     & 5e-5  & 0    & 30  \\ 
\textbf{GigaSpeech}~\cite{chen2021gigaspeech} & 200     & 1e-5  & 0    & 40  \\ 
\textbf{LibriSpeech}~\cite{panayotov2015librispeech} & 100     & 1e-5  & 0.15 & 40  \\ 
\textbf{TED-LIUM}~\cite{hernandez2018ted}    & 200     & 5e-5  & 0    & 40  \\
\textbf{VP}~\cite{wang-etal-2021-voxpopuli}  & 100     & 5e-5  & 0    & 35  \\\bottomrule
\end{tabular}}
\caption{Hyper-parameters for the English ASR models following the setup from \citet{10446931}.}
\label{tab:appendix:mono:hyper}
\end{table}

\begin{table}
\centering
\resizebox{\columnwidth}{!}{
\begin{tabular}{ccc}\toprule
             & HuBERT-base & \mmhubert{} \\\midrule
\textbf{AMI}          & \textbf{33.1} & 33.6 (+0.5) \\
\textbf{CV}           & 37.2 & \textbf{35.4} (-1.8) \\
\textbf{Earnings-22}  & 35.3 & \textbf{34.7} (-0.6) \\
\textbf{GigaSpeech}   & \textbf{25.9} & 26.3 (+0.4) \\
\textbf{LibriSpeech-clean}  & \textbf{7.7} & 9.7 (+2.0) \\
\textbf{LibriSpeech-other}  & \textbf{13.6} & 17.3 (+3.7) \\
\textbf{TED-LIUM}     & \textbf{11.7} & 13.1 (+1.4) \\
\textbf{VP}    & \textbf{18.7} & 19.0 (+0.3) \\\midrule
\textbf{Average WER ($\downarrow$)} & \textbf{23.1} & 23.6 (+0.5) \\\bottomrule
\end{tabular}}
\caption{WER ($\downarrow$) scores for English ASR systems trained on the different datasets from the ESB Benchmark. Following~\citet{10446931}, only 50\,h of training data are used. The score difference between \mmhubert{} and HuBERT-base scores is presented between parentheses.}
\label{tab:appendix:monolingual:multilingual}
\end{table}

\clearpage
\section{Data}
\label{app.data}

\subsection{Validation Set}
The validation set is the same for all our experiments. It was created by sampling 5 utterances per (language, source) pair from the training set, resulting in 1,275 examples. These examples are not removed from the training set.

\subsection{VoxLingua107 Dataset Filtering}\label{app.data.voxlingua}

Manual inspection revealed that some language splits of this dataset contained a considerable amount of music-, noise- and silence-only files. These occurrences were more frequent in less-resourced languages. In order to increase the quality of the data we feed into our self-supervised models, and to more accurately estimate the amount of speech data per language they learn from, we filtered this dataset by performing music, noise and silence detection using the \texttt{inaSpeechSegmenter} tool~\citep{ddoukhanicassp2018}. 

Using this tool's default settings, and knowing that the average utterance length for this dataset is $9s$, we classify a file as \textit{music} if a music event is detected for longer than $2s$. Similarly, a file is classified as \textit{noise} if there is a noise event for longer than $2s$, or if a \textit{non-energy} event~(i.e. silence) is detected for longer than $5s$. These settings were optimized using Cebuano, a language for which we observed a significant amount of noisy utterances. In total, we removed over 332K potentially noisy utterances~(249K music, 83K noise/silence). We make both our filtering script\footnote{Scripts available at: \url{https://github.com/utter-project/mHuBERT-147-scripts/tree/main/01_dataset_preproc/dataset_specific_preproc/voxlingua107}} and list of retained VL files available.\footnote{Manifest files available at: \url{https://huggingface.co/utter-project/mHuBERT-147-base-3rd-iter/tree/main/manifest}}

Table~\ref{tab:data:vlvad} presents the 107 languages of this dataset sorted into buckets corresponding to the percentage of noisy input detected by the automatic tool. We notice that for most languages, this percentage is limited: 77 languages have noise below 15\%. However, for some low-resource languages, the amount of identified noisy utterances can go up to 80\%. We take these automatic filtering results cautiously -- as they might not be truly representative of the amount of noise present in \textit{all} of these languages. However, we do believe they illustrate an existing quality issue of this popular corpus for some languages.

\begin{table}[]
\centering
\scriptsize
\begin{tabular}{ccl}
\toprule
\textbf{\% of Noise} & \textbf{Count} & \textbf{Languages} \\\midrule
$[0,5]$                        & 11                   & abk, hye, kat, lit, ltz, sqi, tuk                                                                                                                                                                                                                                                                    \\\midrule
(5,10]                       & 44                   & \begin{tabular}[c]{@{}l@{}}
afr, amh, ara, asm, aze, bak, bel, \\
bod, bos, bul, cat, ces, deu, ell, \\
epo, est, fas, fra, glg, heb, hrv, \\
hun, ita, kaz, kor, mkd, mlt, mon, \\
nep, nld, nor, pol, por, pus, ron, \\
rus, sin, slk, slv, som, srp, swe, \\
tat, tgk, tur, ukr, uzb, zh-CN\end{tabular} \\\midrule
(10,15]                      & 22                   & \begin{tabular}[c]{@{}l@{}}
dan, eng, fil, fin, hat, hau, hin, \\
ind, isl, jpn, kan, lav, lin, mal, \\
mar, mlg, msa, oci, spa, swa, \\tam, vie\end{tabular}\\\midrule  
(15,20]                      & 10                   & 

\begin{tabular}[c]{@{}l@{}}
ben, cym, eus, fao, glv, \\guj, lao, tel, tha, urd      \end{tabular}\\\midrule
(20,30]                      & 6                    & bre, ina, mya, nno, pan, yid                                                                                                                                                                                                                                                                         \\\midrule
(30,40]                      & 3                    & sna, sun, yor                                                                                                                                                                                                                                                                                        \\\midrule
(40,60]                      & 6                    & ceb, jav, khm, mri, sco, snd                                                                                                                                                                                                                                                                         \\\midrule
(50,80]                      & 5                    & grn, haw, lat, san, war\\\bottomrule           
\end{tabular}
\caption{The overall percentage of noise~(music, silence, noise) detected in the different language splits of VL using an automatic tool.}
\label{tab:data:vlvad}
\end{table}

Lastly, we note that another key issue in this corpus is the miscategorization of language in many speech utterances. In particular, we observed many English utterances that were labelled as belonging to other languages, with this issue being far more prevalent for low-resource languages. Since resolving this would require a powerful language identification tool to work well in all of the 107 languages present in this corpus, we decided to not filter or remove these utterances -- but we do highlight that the number of utterances in low-resourced languages for this corpus is probably an over-estimation.

%
\begin{table*}
\resizebox{\textwidth}{!}{
\begin{tabular}{lll}
\toprule
\textbf{Dataset}& \textbf{Full Name} & \textbf{Download URL}\\\midrule
\textbf{Aishell}& Aishell~\cite{aishell_2017} & \url{https://www.openslr.org/33/}\\\hline
\textbf{Aishell-3}& AISHELL-3~\cite{AISHELL-3_2020} &\url{https://www.openslr.org/93/}\\\hline
\textbf{B-TTS}& BibleTTS~\cite{meyer2022bibletts} &\url{http://www.openslr.org/129/}\\\hline
\textbf{Clovacall}& ClovaCall~\cite{ha20_interspeech} & \url{https://github.com/clovaai/ClovaCall}\\\hline
\textbf{CV}& Common Voice version 11.0~\cite{commonvoice:2020}&\url{https://commonvoice.mozilla.org/en/datasets}\\\hline
\multirow{6}{*}{\textbf{G-TTS}}& High quality TTS data for Javanese~\cite{kjartansson-etal-tts-sltu2018}& \url{http://www.openslr.org/41/}\\
& High quality TTS data for Khmer~\cite{kjartansson-etal-tts-sltu2018}& \url{http://www.openslr.org/42/}\\
& High quality TTS data for Nepali~\cite{kjartansson-etal-tts-sltu2018}& \url{http://www.openslr.org/43/}\\
& High quality TTS data for Sundanese~\cite{kjartansson-etal-tts-sltu2018}& \url{http://www.openslr.org/44/}\\
& High quality TTS data for four South African languages~\cite{van-niekerk-etal-2017} & \url{http://www.openslr.org/32/}\\
& High quality TTS data for Bengali languages~\cite{kjartansson-etal-tts-sltu2018}& \url{https://www.openslr.org/37/}\\\hline
\multirow{2}{*}{\textbf{IISc-MILE}} & IISc-MILE Tamil ASR Corpus~\cite{mile_1,mile_2}& \url{https://www.openslr.org/127/}\\
& IISc-MILE Kannada ASR Corpus~\cite{mile_1,mile_2}& \url{http://www.openslr.org/126/}\\\hline
\textbf{JVS}& Japanese versatile speech~\cite{takamichi2019jvs}       &\url{https://sites.google.com/site/shinnosuketakamichi/research-topics/jvs_corpus} \\\hline
\textbf{Kokoro}& Kokoro& \url{https://github.com/kaiidams/Kokoro-Speech-Dataset}\\\hline
\textbf{kosp2e}& Korean Speech to English Translation Corpus~\cite{cho21b_interspeech}& \url{https://github.com/warnikchow/kosp2e}\\\hline
\textbf{MLS}& Multilingual LibriSpeech~\cite{Pratap2020MLSAL}& \url{http://www.openslr.org/94/}\\\hline
\textbf{MS}& MediaSpeech~\cite{mediaspeech2021}& \url{https://www.openslr.org/108/}\\\hline
\textbf{Samrómur}& Samrómur Unverified 22.07~\cite{hedstrom2022}& \url{https://www.openslr.org/128/}\\\hline
\textbf{THCHS-30}& THCHS-30~\cite{THCHS30_2015}& \url{https://www.openslr.org/18/}\\\hline
\textbf{THUYG-20}& THUYG-20~\cite{THUGY20_2015}& \url{https://www.openslr.org/22/}\\\hline
\textbf{VL}& VoxLingua107~\cite{valk2021slt}& \url{https://bark.phon.ioc.ee/voxlingua107/}\\\hline
\textbf{VP}& VoxPopuli~\cite{wang-etal-2021-voxpopuli}&  \url{https://github.com/facebookresearch/voxpopuli/}\\\bottomrule
\end{tabular}}
\caption{Downloading URLs for \mmhubert{} data. All data was downloaded between November 2022 and February 2023.}
\label{tab:url}
\end{table*}

%
\begin{table}[]
\centering
\scriptsize
\begin{tabular}{cccc}
\toprule
\textbf{IETF code}   & \textbf{Language}            & \textbf{Dataset} & \textbf{\# Hours} \\\midrule
\multirow{2}{*}{abk} & \multirow{2}{*}{Abkhazian}   & CV               & 34.0             \\
                     &                              & VL               & 10.0             \\\midrule
\multirow{2}{*}{afr} & \multirow{2}{*}{Afrikaans}   & G-TTS            & 3.3              \\
                     &                              & VL               & 100.0            \\\midrule
amh                  & Amharic                      & VL               & 74.0             \\\midrule
\multirow{2}{*}{ara} & \multirow{2}{*}{Arabic}      & CV               & 61.0             \\
                     &                              & VL               & 53.6             \\\midrule
\multirow{2}{*}{asm} & \multirow{2}{*}{Assamese}    & CV               & 1.0              \\
                     &                              & VL               & 143.0            \\\midrule
ast                  & Asturian                     & CV               & 0.01              \\\midrule
\multirow{2}{*}{aze} & \multirow{2}{*}{Azerbaijani} & CV               & 0.05              \\
                     &                              & VL               & 55.6             \\\midrule
\multirow{2}{*}{bak} & \multirow{2}{*}{Bashkir}     & CV               & 213.4            \\
                     &                              & VL               & 54.7             \\\midrule
bas                  & Basaa                        & CV               & 0.9              \\\midrule
\multirow{2}{*}{bel} & \multirow{2}{*}{Belarusian}  & CV               & 1,101.6           \\
                     &                              & VL               & 126.4            \\\midrule
\multirow{3}{*}{ben} & \multirow{3}{*}{Bengali}     & CV               & 33.2             \\
                     &                              & G-TTS            & 2.0              \\
                     &                              & VL               & 45.9             \\\midrule
bod                  & Tibetan                      & VL               & 92.4             \\\midrule
bos                  & Bosnian                      & VL               & 97.7             \\\midrule
\multirow{2}{*}{bre} & \multirow{2}{*}{Breton}      & CV               & 4.4              \\
                     &                              & VL               & 33.1             \\\midrule
\multirow{3}{*}{bul} & \multirow{3}{*}{Bulgarian}   & CV               & 4.6              \\
                     &                              & VL               & 47.7             \\
                     &                              & VP               & 773.3            \\\midrule
\multirow{2}{*}{cat} & \multirow{2}{*}{Catalan}     & CV               & 1,638.7           \\
                     &                              & VL               & 80.3             \\\midrule
ceb                  & Cebuano                      & VL               & 3.9              \\\midrule
\multirow{3}{*}{ces} & \multirow{3}{*}{Czech}       & CV               & 38.2             \\
                     &                              & VL               & 62.4             \\
                     &                              & VP               & 866.1            \\\midrule
chv                  & Chuvash                      & CV               & 17.3             \\\midrule
ckb                  & Central Kurdish              & CV               & 90.5             \\\midrule
cnh                  & Hakha Chin                   & CV               & 0.7              \\\midrule
\multirow{2}{*}{cym} & \multirow{2}{*}{Welsh}       & CV               & 100.6            \\
                     &                              & VL               & 65.6             \\\midrule
\multirow{3}{*}{dan} & \multirow{3}{*}{Danish}      & CV               & 3.5              \\
                     &                              & VL               & 25.0             \\
                     &                              & VP               & 728.4            \\\midrule
\multirow{4}{*}{deu} & \multirow{4}{*}{German}      & CV               & 1,091.0           \\
                     &                              & MLS              & 1,966.5           \\
                     &                              & VL               & 36.3             \\
                     &                              & VP               & 274.8            \\\midrule
div                  & Maldivian                    & CV               & 32.4             \\\midrule
\multirow{3}{*}{ell} & \multirow{3}{*}{Greek}       & CV               & 13.0             \\
                     &                              & VL               & 60.5             \\
                     &                              & VP               & 573.2            \\\midrule
\multirow{4}{*}{eng} & \multirow{4}{*}{English}     & CV               & 2,213.9           \\
                     &                              & MLS              & 44,659.7           \\
                     &                              & VL               & 43.5             \\
                     &                              & VP               & 293.6            \\\midrule
\multirow{2}{*}{epo} & \multirow{2}{*}{Esperanto}   & CV               & 1,355.6           \\
                     &                              & VL               & 8.9              \\\midrule
\multirow{3}{*}{est} & \multirow{3}{*}{Estonian}    & CV               & 29.9             \\
                     &                              & VL               & 34.7             \\
                     &                              & VP               & 682.7            \\
\bottomrule            
\end{tabular}                
\caption{\textbf{(1/5)} List of included languages, with corresponding amount of speech data per dataset.}
\label{tab:languagefull:part1}
\end{table}
%
\begin{table}[]
\centering
\scriptsize
\begin{tabular}{cccc}
\toprule
\textbf{IETF code}   & \textbf{Language}            & \textbf{Dataset} & \textbf{\# Hours} \\\midrule
\multirow{2}{*}{eus} & \multirow{2}{*}{Basque}      & CV               & 79.0             \\
                     &                              & VL               & 25.5             \\\midrule
ewe                  & Ewe                          & B-TTS            & 76.6  
\\\midrule
fao                  & Faroese                      & VL               & 56.4             \\\midrule
\multirow{2}{*}{fas} & \multirow{2}{*}{Persian}     & CV               & 312.4            \\
                     &                              & VL               & 52.0   \\\midrule
fil                  & Tagalog                      & VL               & 82.7             \\\midrule
\multirow{3}{*}{fin} & \multirow{3}{*}{Finnish}     & CV               & 5.0              \\
                     &                              & VL               & 29.5             \\
                     &                              & VP               & 816.1            \\\midrule
\multirow{4}{*}{fra} & \multirow{4}{*}{French}      & CV               & 836.0            \\
                     &                              & MLS              & 1,076.6           \\
                     &                              & VL               & 63.2             \\
                     &                              & VP               & 290.8            \\\midrule
fry                  & Frisian                      & CV               & 41.2             \\\midrule
gle                  & Irish                        & CV               & 3.2              \\\midrule
\multirow{2}{*}{glg} & \multirow{2}{*}{Galician}    & CV               & 4.4              \\
                     &                              & VL               & 66.0             \\\midrule
glv                  & Manx Gaelic                  & VL               & 3.5              \\\midrule
\multirow{2}{*}{grn} & \multirow{2}{*}{Guarani}     & CV               & 0.4              \\
                     &                              & VL               & 1.0              \\\midrule
guj                  & Gujarati                     & VL               & 39.5             \\\midrule
hat                  & Haitian Creole               & VL               & 86.3             \\\midrule
\multirow{3}{*}{hau} & \multirow{3}{*}{Hausa}       & B-TTS            & 85.6             \\
                     &                              & CV               & 2.3              \\
                     &                              & VL               & 81.1             \\\midrule
haw                  & Hawaiian                     & VL               & 5.4              \\\midrule
heb                  & Hebrew                       & VL               & 89.7             \\\midrule
\multirow{2}{*}{hin} & \multirow{2}{*}{Hindi}       & CV               & 5.0              \\
                     &                              & VL               & 73.2             \\\midrule
\multirow{2}{*}{hrv} & \multirow{2}{*}{Croatian}    & VL               & 109.4            \\
                     &                              & VP               & 836.6            \\\midrule
hsb                  & Upper Sorbian                & CV               & 1.5              \\\midrule
\multirow{3}{*}{hun} & \multirow{3}{*}{Hungarian}   & CV               & 9.5              \\
                     &                              & VL               & 68.8             \\
                     &                              & VP               & 851.0            \\\midrule
\multirow{2}{*}{hye} & \multirow{2}{*}{Armenian}    & CV               & 1.0              \\
                     &                              & VL               & 66.5             \\\midrule
ibo                  & Igbo                         & CV               & 0.01              \\\midrule
\multirow{2}{*}{ina} & \multirow{2}{*}{Interlingua} & CV               & 8.3              \\
                     &                              & VL               & 2.1              \\\midrule
\multirow{2}{*}{ind} & \multirow{2}{*}{Indonesian}  & CV               & 20.6             \\
                     &                              & VL               & 34.4             \\\midrule
\multirow{2}{*}{isl} & \multirow{2}{*}{Icelandic}   & samromur         & 2,088.3           \\
                     &                              & VL               & 81.2             \\\midrule
\multirow{4}{*}{ita} & \multirow{4}{*}{Italian}     & CV               & 271.9            \\
                     &                              & MLS              & 247.4            \\
                     &                              & VL               & 46.6             \\
                     &                              & VP               & 613.8            \\\midrule
\multirow{2}{*}{jav} & \multirow{2}{*}{Javanese}    & G-TTS            & 3.5              \\
                     &                              & VL               & 24.4             \\\midrule
\multirow{4}{*}{jpn} & \multirow{4}{*}{Japanese}    & CV               & 37.0             \\
                     &                              & JVS              & 26.4             \\
                     &                              & kokoro           & 60.0             \\
                     &                              & VL               & 50.1             \\\midrule
kab                  & Kabyle                       & CV               & 516.6            \\
\bottomrule            
\end{tabular}                
\caption{\textbf{(2/5)} List of included languages, with corresponding amount of speech data per dataset.}
\label{tab:languagefull:part2}
\end{table}
%
\begin{table}[]
\centering
\scriptsize
\begin{tabular}{cccc}
\toprule
\textbf{IETF code}   & \textbf{Language}            & \textbf{Dataset} & \textbf{\# Hours} \\\midrule
\multirow{2}{*}{kan} & \multirow{2}{*}{Kannada}     & IISc-MILE        & 273.8            \\
                     &                              & VL               & 39.9             \\\midrule
\multirow{2}{*}{kat} & \multirow{2}{*}{Georgian}    & CV               & 6.6              \\
                     &                              & VL               & 93.4             \\\midrule
\multirow{2}{*}{kaz} & \multirow{2}{*}{Kazakh}      & CV               & 0.6              \\
                     &                              & VL               & 72.6             \\\midrule
\multirow{2}{*}{khm} & \multirow{2}{*}{Khmer}       & G-TTS            & 4.0              \\
                     &                              & VL               & 27.0             \\\midrule
kin                  & Kinyarwanda                  & CV               & 1,982.7           \\\midrule
kir                  & Kyrgyz                       & CV               & 32.6             \\\midrule
kmr                  & Northern Kurdish             & CV               & 45.8             \\\midrule
\multirow{3}{*}{kor} & \multirow{3}{*}{Korean}            & clovacall        & 38.1             \\
                     &                                    & kosp2e           & 190.9            \\
                     &                                    & VL               & 71.5             \\\midrule
lao                  & Lao                                & VL               & 36.1             \\\midrule
lat                  & Latin                              & VL               & 30.9             \\\midrule
\multirow{3}{*}{lav} & \multirow{3}{*}{Latvian}           & CV               & 3.0              \\
                     &                                    & VL               & 37.0             \\
    &              & VP               & 868.4            \\\midrule
\multirow{2}{*}{lin} & \multirow{2}{*}{Lingala}           & B-TTS            & 54.0             \\
                     &                                    & VL               & 77.8             \\\midrule
\multirow{3}{*}{lit} & \multirow{3}{*}{Lithuanian}        & CV               & 7.1              \\
                     &                                    & VL               & 78.4             \\
                     &                                    & VP               & 796.2            \\\midrule
ltz                  & Luxembourgish                      & VL               & 71.8             \\\midrule
lug                  & Ganda                              & CV               & 363.0            \\\midrule
\multirow{2}{*}{mal} & \multirow{2}{*}{Malayalam}         & CV               & 0.5              \\
                     &                                    & VL               & 42.5             \\\midrule
\multirow{2}{*}{mar} & \multirow{2}{*}{Marathi}           & CV               & 11.9             \\
                     &                                    & VL               & 74.7             \\\midrule
mdf                  & Moksha                             & CV               & 0.2              \\\midrule
mhr                  & Meadow Mari                       & CV               & 97.5             \\\midrule
\multirow{2}{*}{mkd} & \multirow{2}{*}{Macedonian}        & CV               & 0.2              \\
                     &                                    & VL               & 103.9            \\\midrule
mlg                  & Malagasy                           & VL               & 94.5             \\\midrule
\multirow{3}{*}{mlt} & \multirow{3}{*}{Maltese}           & CV               & 3.9              \\
                     &                                    & VL               & 62.9             \\
                     &                                    & VP               & 818.1            \\\midrule
\multirow{2}{*}{mon} & \multirow{2}{*}{Mongolian}         & CV               & 6.6              \\
                     &                                    & VL               & 65.9             \\\midrule
mri                  & Māori                              & VL               & 20.5             \\\midrule
mrj                  & Western Mari                       & CV               & 6.5              \\\midrule
msa                  & Malay                              & VL               & 71.8             \\\midrule
mya                  & Burmese                            & VL               & 31.5             \\\midrule
myv                  & Erzya                              & CV               & 1.1              \\\midrule
nan-tw               & Taiwanese Hokkien                  & CV               & 0.7              \\\midrule
\multirow{3}{*}{nep} & \multirow{3}{*}{Nepali}            & CV               & 0.01              \\
                     &                                    & G-TTS            & 2.8              \\
                     &                                    & VL               & 65.6             \\\midrule
\multirow{4}{*}{nld} & \multirow{4}{*}{Dutch}             & CV               & 72.1             \\
                     &                                    & MLS              & 1,554.2           \\
                     &                                    & VL               & 37.2             \\
                     &                                    & VP               & 277.0            \\\midrule
\multirow{2}{*}{nno} & \multirow{2}{*}{Norwegian Nynorsk} & CV               & 0.4              \\
                     &                                    & VL               & 43.3             \\\midrule
nor                  & Norwegian                          & VL               & 98.2             \\\midrule
oci                  & Occitan                            & VL               & 13.7             \\
\bottomrule            
\end{tabular}                
\caption{\textbf{(3/5)} List of included languages, with corresponding amount of speech data per dataset.}
\label{tab:languagefull:part3}
\end{table}
%
\begin{table}[]
\centering
\scriptsize
\begin{tabular}{cccc}
\toprule
\textbf{IETF code}   & \textbf{Language}            & \textbf{Dataset} & \textbf{\# Hours} \\\midrule
ori                  & Odia                               & CV               & 0.8              \\\midrule
\multirow{2}{*}{pan} & \multirow{2}{*}{Punjabi}           & CV               & 1.0              \\
                     &                                    & VL               & 42.1             \\\midrule
\multirow{4}{*}{pol} & \multirow{4}{*}{Polish}            & CV               & 129.4            \\
                     &                                    & MLS              & 103.6            \\
                     &                                    & VL               & 76.2             \\
                     &                                    & VP               & 841.9            \\\midrule
\multirow{4}{*}{por} & \multirow{4}{*}{Portuguese}        & CV               & 102.3            \\
                     &                                    & MLS              & 161.0            \\
                     &                                    & VL               & 58.6             \\
                     &                                    & VP               & 851.6            \\\midrule
pus                  & Pashto                             & VL               & 44.7             \\\midrule
rm-sursilv           & Sursilvan                          & CV               & 2.4              \\\midrule
rm-vallader          & Vallader                           & CV               & 1.2       \\\midrule
\multirow{3}{*}{ron} & \multirow{3}{*}{Romanian}  & CV               & 8.5              \\
                     &                            & VL               & 59.0             \\
                     &                            & VP               & 834.8            \\\midrule
\multirow{2}{*}{rus} & \multirow{2}{*}{Russian}   & CV               & 149.1            \\
                     &                            & VL               & 67.5             \\\midrule
sah                  & Sakha                      & CV               & 2.7              \\\midrule
san                  & Sanskrit                   & VL               & 4.6              \\\midrule
sat                  & Santali                    & CV               & 0.4              \\\midrule
sco                  & Scots                      & VL               & 1.5              \\\midrule
sin                  & Sinhala                    & VL               & 60.6             \\\midrule
skr                  & Saraiki                    & CV               & 1.3              \\\midrule
\multirow{3}{*}{slk} & \multirow{3}{*}{Slovak}    & CV               & 12.9             \\
                     &                            & VL               & 36.6             \\
                     &                            & VP               & 644.5            \\\midrule
\multirow{3}{*}{slv} & \multirow{3}{*}{Slovenian} & CV               & 7.6              \\
                     &                            & VL               & 112.3            \\
                     &                            & VP               & 832.1            \\\midrule
sna                  & Shona                      & VL               & 21.5             \\\midrule
snd                  & Sindhi                     & VL               & 48.2             \\\midrule
som                  & Somali                     & VL               & 94.9             \\\midrule
sot                  & Southern Sotho             & G-TTS            & 3.2              \\\midrule
\multirow{4}{*}{spa} & \multirow{4}{*}{Spanish}   & CV               & 380.7            \\
                     &                            & MLS              & 917.7            \\
                     &                            & VL               & 33.9             \\
                     &                            & VP               & 301.5            \\\midrule
sqi                  & Albanian                   & VL               & 67.2             \\\midrule
srd                  & Sardinian                  & CV               & 0.7              \\\midrule
\multirow{2}{*}{srp} & \multirow{2}{*}{Serbian}   & CV               & 0.8              \\
                     &                            & VL               & 47.5             \\\midrule
\multirow{2}{*}{sun} & \multirow{2}{*}{Sundanese} & G-TTS            & 2.1              \\
                     &                            & VL               & 43.6             \\\midrule
\multirow{2}{*}{swa} & \multirow{2}{*}{Swahili}   & CV               & 304.2            \\
                     &                            & VL               & 57.6             \\\midrule
\multirow{3}{*}{swe} & \multirow{3}{*}{Swedish}   & CV               & 29.5             \\
                     &                            & VL               & 31.3             \\
                     &                            & VP               & 827.4            \\\midrule
                     \multirow{3}{*}{tam} & \multirow{3}{*}{Tamil}     & CV               & 186.1            \\
                     &                            & IISc-MILE        & 132.2            \\
                     &                            & VL               & 44.1             \\\midrule
\multirow{2}{*}{tat} & \multirow{2}{*}{Tatar}     & CV               & 20.3             \\
                     &                            & VL               & 93.4             \\\midrule
tel                  & Telugu                     & VL               & 64.3             \\\midrule
tgk                  & Tajik                      & VL               & 60.5             \\\bottomrule            
\end{tabular}                
\caption{\textbf{(4/5)} List of included languages, with corresponding amount of speech data per dataset.}
\label{tab:languagefull:part4}
\end{table}
%
\begin{table}[]
\centering
\scriptsize
\begin{tabular}{cccc}
\toprule
\textbf{IETF code}   & \textbf{Language}            & \textbf{Dataset} & \textbf{\# Hours} \\\midrule
\multirow{2}{*}{tha} & \multirow{2}{*}{Thai}      & CV               & 134.7            \\
                     &                            & VL               & 50.8             \\\midrule
tig                  & Tigre                      & CV               & 0.01              \\\midrule
tpi                  & Tok Pisin                  & CV               & 3.3              \\\midrule
tsn                  & Tswana                     & G-TTS            & 3.5              \\\midrule
tuk                  & Turkmen                    & VL               & 81.8             \\\midrule
\multirow{3}{*}{tur} & \multirow{3}{*}{Turkish}   & CV               & 59.3             \\
                     &                            & MS               & 10.0             \\
                     &                            & VL               & 54.5             \\\midrule
tw-akuapem           & Akwapem Twi                & B-TTS            & 59.8             \\\midrule
tw-asante            & Asante Twi                 & B-TTS            & 56.8             \\\midrule
uig                  & Uyghur                     & CV               & 94.9             \\\midrule
uig                  & Uyghur                     & THUYG-20         & 20.7             \\\midrule
\multirow{2}{*}{ukr} & \multirow{2}{*}{Ukrainian} & CV               & 52.5             \\
                     &                            & VL               & 49.7             \\\midrule
\multirow{2}{*}{urd} & \multirow{2}{*}{Urdu}      & CV               & 38.8             \\
                     &                            & VL               & 35.2      \\\midrule
\multirow{2}{*}{uzb}   & \multirow{2}{*}{Uzbek}         & CV               & 69.7             \\
                       &                                & VL               & 42.2             \\\midrule
\multirow{2}{*}{vie}   & \multirow{2}{*}{Vietnamese}    & CV               & 3.8              \\
                       &                                & VL               & 55.5             \\\midrule
vot                    & Votic                          & CV               & 0.1              \\\midrule
war                    & Waray                          & VL               & 3.9              \\\midrule
xho                    & Xhosa                          & G-TTS            & 3.1              \\\midrule
yid                    & Yiddish                        & VL               & 35.9             \\\midrule
\multirow{2}{*}{yor}   & \multirow{2}{*}{Yoruba}        & B-TTS            & 24.8             \\
                       &                                & VL               & 66.0             \\\midrule
yue                    & Cantonese                      & CV               & 15.6             \\\midrule
\multirow{5}{*}{zh-CN} & \multirow{5}{*}{Chinese (PRC)} & Aishell          & 151.1            \\
                       &                                & Aishell-3        & 60.6             \\
                       &                                & CV               & 104.6            \\
                       &                                & THCHS-30         & 25.5             \\
                       &                                & VL               & 40.9             \\\midrule
zh-HK                  & Chinese (Hong Kong)            & CV               & 89.5             \\\midrule
zh-TW                  & Chinese (Taiwan)               & CV               & 56.3        \\\midrule
\multicolumn{3}{c}{\textbf{Total}} & \textbf{90,429.5} \\
\bottomrule            
\end{tabular}                
\caption{\textbf{(5/5)} List of included languages, with corresponding amount of speech data per dataset.}
\label{tab:languagefull:part5}
\end{table}

\end{document}